\documentclass{article}
\pdfoutput=1

\usepackage[preprint, nonatbib]{neurips_data_2023}

\usepackage[utf8]{inputenc} 
\usepackage[T1]{fontenc}    
\usepackage{url}            
\usepackage{booktabs}       
\usepackage{amsfonts}       
\usepackage{nicefrac}       
\usepackage{microtype}      
\usepackage{xcolor}         
\usepackage{tabu}         
\usepackage{xspace,mfirstuc,tabulary}

\usepackage{multirow}
\usepackage{array, caption, floatrow, makecell, booktabs}
\usepackage{wrapfig}
\usepackage{floatrow}
\usepackage{comment}
\usepackage{footnote}
\usepackage{enumitem}
\usepackage{accents}

\usepackage{graphicx}
\usepackage{bbding}
\usepackage{ntheorem}
\usepackage{enumitem}
\usepackage{oplotsymbl}
\usepackage{multirow}
\usepackage{bbm}
\usepackage{adjustbox}
\usepackage{subcaption}
\usepackage{caption}
\usepackage{multicol}
\usepackage{wrapfig}

\usepackage{wrapfig}
\usepackage{comment}
\usepackage{enumitem}
\usepackage{accents}

\usepackage{tabu}

\usepackage{multirow,array}
\usepackage{color, colortbl}
\definecolor{Gray}{gray}{0.93}

\usepackage{booktabs}
\usepackage{pifont}

\usepackage[pagebackref=true,breaklinks=true,colorlinks,bookmarks=false]{hyperref}

\usepackage{floatrow}
\newlength\savewidth
\newcommand\paperurl[1]{{\footnotesize{\color{blue}{\url{#1}}}}}

\definecolor{rose}{HTML}{003472}

\newcommand{\sota}[1]{\textcolor{black}{$\textbf{#1}$}}

\title{Grounding DINO 1.5: Advance the ``Edge'' of Open-Set Object Detection}

\author{%
 \textbf{Tianhe Ren\thanks{Equal contributions. List order is random.}\;,\;  Qing Jiang$^{\mathrm{*}}$,\; Shilong Liu$^{\mathrm{*}}$,\; Zhaoyang Zeng$^{\mathrm{*}}$,\;}\\
  \textbf{ Wenlong Liu,\; Han Gao,\; Hongjie Huang,\; Zhengyu Ma,\; Xiaoke Jiang,\;}\\
  \textbf{Yihao Chen,\; Yuda Xiong,\; Hao Zhang,\; Feng Li,\; Peijun Tang,\; Kent Yu,\; Lei Zhang\thanks{Project lead and corresponding author.}}\\
  \\
  International Digital Economy Academy (IDEA), IDEA Research \\
  \url{https://deepdataspace.com/home}
}

\begin{document}

\maketitle

\begin{abstract}



This paper introduces Grounding DINO 1.5, a suite of advanced open-set object detection models developed by IDEA Research, which aims to advance the “Edge”{\footnotemark[1]} of open-set object detection.
The suite encompasses two models: \textit{Grounding DINO 1.5 Pro}, a high-performance model designed for \emph{stronger} generalization capability across a wide range of scenarios, and \textit{Grounding DINO 1.5 Edge}, an efficient model optimized for \emph{faster} speed demanded in many applications requiring edge deployment.
The \textit{Grounding DINO 1.5 Pro} model advances its predecessor by scaling up the model architecture, integrating an enhanced vision backbone, and expanding the training dataset to over 20 million images with grounding annotations, thereby achieving a richer semantic understanding.
The \textit{Grounding DINO 1.5 Edge} model, while designed for efficiency with reduced feature scales, maintains robust detection capabilities by being trained on the same comprehensive dataset.
Empirical results demonstrate the effectiveness of Grounding DINO 1.5, with the \textit{Grounding DINO 1.5 Pro} model attaining a 54.3 AP on the COCO detection benchmark and a 55.7 AP on the LVIS-minival zero-shot transfer benchmark, setting new records for open-set object detection. Furthermore, the \textit{Grounding DINO 1.5 Edge} model, when optimized with TensorRT, achieves a speed of 75.2 FPS while attaining a zero-shot performance of 36.2 AP on the LVIS-minival benchmark, making it more suitable for edge computing scenarios. Model examples and demos with API will be released at \href{https://github.com/IDEA-Research/Grounding-DINO-1.5-API}{https://github.com/IDEA-Research/Grounding-DINO-1.5-API}.

\footnotetext[1]{We use ``edge'' for its dual meaning both as in pushing the boundaries and as in running on edge devices.}

\end{abstract}

\section{Introduction}
In this paper, we introduce Grounding DINO 1.5, a series of powerful and practical open-set object detection models developed by IDEA Research. 
Object detection is a fundamental task in computer vision, with recent efforts focusing on developing generic detectors capable of performing detection across a wide variety of real-world applications. 
A key strategy for improving model generalization across diverse object categories is the integration of language modality, which has received increasingly more attention and has undergone extensive development in recent research.

Grounding DINO~\cite{GroundingDINO} represents a significant advancement in this area. 
Building on the Transformer-based DINO~\cite{DINO} architecture, it incorporates linguistic information to enable open-set object detection in various scenarios. 
Following GLIP~\cite{GLIP}, Grounding DINO redefines object detection as a phrase grounding task, facilitating large-scale pre-training using both detection and grounding datasets. 
This approach, coupled with self-training on pseudo-labeled grounding data derived from an almost unlimited pool of image-text pairs, enhances the model’s applicability to open-world settings due to its robust architecture and semantic-rich dataset.


Building upon the success of Grounding DINO, Grounding DINO 1.5 extends the model's capabilities in two key areas: \emph{stronger} detection performance and \emph{faster} inference speed, designated as the Grounding DINO 1.5 Pro and Grounding DINO 1.5 Edge models, respectively.

The Grounding DINO 1.5 Pro model significantly expands both the model's capacity and the dataset size, with the goal of creating a more potent and versatile open-set object detection model. Specifically, we have upscaled the model by incorporating the pre-trained ViT-L~\cite{EVA-02} architecture and developed a data engine capable of producing over 20 million images with grounding annotations from diverse sources, thereby enriching the model's semantic comprehension. 

By contrast, the Grounding DINO 1.5 Edge model is tailored for edge devices, focusing on computational efficiency without compromising detection quality. We develop an efficient feature enhancer that leverages only high-level image features, removing the need of multi-scale features. This streamlined approach maintains the model's strong context-aware detection capabilities, after training on the same 20 million images as used for the Pro model.

Extensive results from our experiments validate the superiority of Grounding DINO 1.5.
Specifically, Grounding DINO 1.5 Pro achieves a 54.3 AP on the COCO detection zero-shot transfer benchmark and simultaneously achieves a 55.7 AP and a 47.6 AP on the LVIS-minival and LVIS-val zero-shot transfer benchmarks, respectively, setting new records on these benchmarks. Moreover, under TensorRT optimization, the Grounding DINO 1.5 Edge model reaches a speed of 75.2 FPS and achieves a zero-shot performance of 36.2 AP on the LVIS-minival benchmark, making it more suitable for edge computing scenarios.

\section{Model Training}

\subsection{Model Architecture}

\begin{figure}[ht!]
\centering
 \includegraphics[width=1.0\textwidth,keepaspectratio]{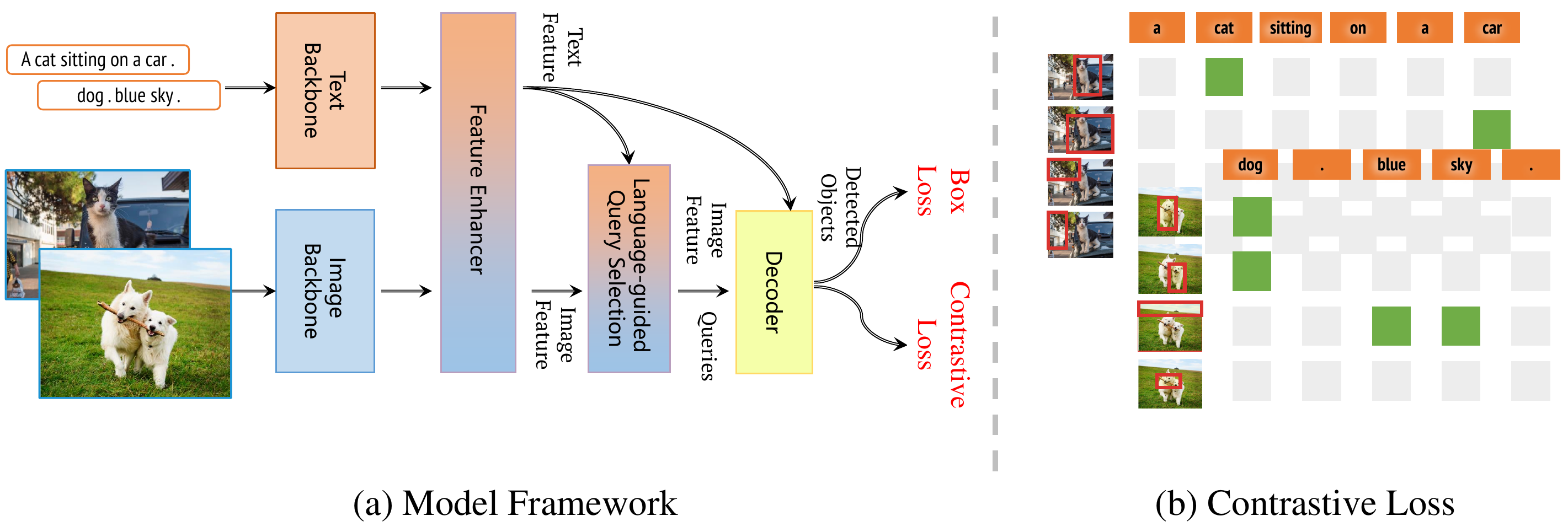}
 \caption{Grounding DINO 1.5 series overall framework.}
 \label{fig:gd1.5_framework}
\end{figure}

We present the overall framework of Grounding DINO 1.5 series in Figure~\ref{fig:gd1.5_framework}. This framework retains the dual-encoder-single-decoder structure of Grounding DINO and extends it into Grounding DINO 1.5 for both the Pro and Edge models, which are introduced in Sections~\ref{gd1.5_pro_introduction} and \ref{gd1.5_edge_introduction}, respectively.

\subsubsection{Grounding DINO 1.5 Pro}
\label{gd1.5_pro_introduction}

Grounding DINO 1.5 Pro preserves the core architecture of Grounding DINO while incorporating a larger Vision Transformer backbone. We adopt the pre-trained ViT-L~\cite{EVA-02} model as our primary vision backbone for its superior performance on downstream tasks and its pure Transformer design, which lays a solid foundation for optimizing the training and inference processes.

Following the methodologies of Grounding DINO~\cite{GroundingDINO} and GLIP~\cite{GLIP}, Grounding DINO 1.5 Pro employs a {deep early fusion} strategy during feature extraction. This involves cross-attention mechanisms between language and image features before the decoding phase, facilitating a more integrated information fusion.

We also compared the strategies of early fusion and later fusion. We observe that models trained with early fusion architecture design tend to yield a higher detection recall and better bounding box precision accuracy. However, this approach can also lead to increased model hallucinations, such as predicting objects not present in the input images.
In contrast to early fusion, the models with a late fusion design, which integrate language and image modalities only in the loss calculation phase, generally demonstrate robustness against hallucinations but may lead to lower detection recall. This is primarily due to the increased challenge of vision and language alignment as late fusion keeps features from different modalities separately until the loss phase. 

Consequently, to simultaneously enhance the model's prediction capability and maintain its robustness during inference, we have retained the early fusion design while introducing a more comprehensive training sampling strategy, which increases the proportion of negative samples during training. Such improvements facilitate achieving a balance between the advantages and drawbacks of the early fusion architecture.

\subsubsection{Grounding DINO 1.5 Edge}
\label{gd1.5_edge_introduction}

Deploying Grounding DINO on edge devices is highly desired by many applications, including autonomous driving, medical image processing, computational photography, etc. However, there is a large gap between the computational cost required by an open-set detection model and the limited resources available on edge devices. 
Grounding DINO uses multi-scale image features and a heavy computational feature enhancer for faster training and better performance, which is impractical for real-time scenarios in real-world applications.

\begin{figure}[ht!]
\centering
\includegraphics[width=0.95\textwidth,keepaspectratio]{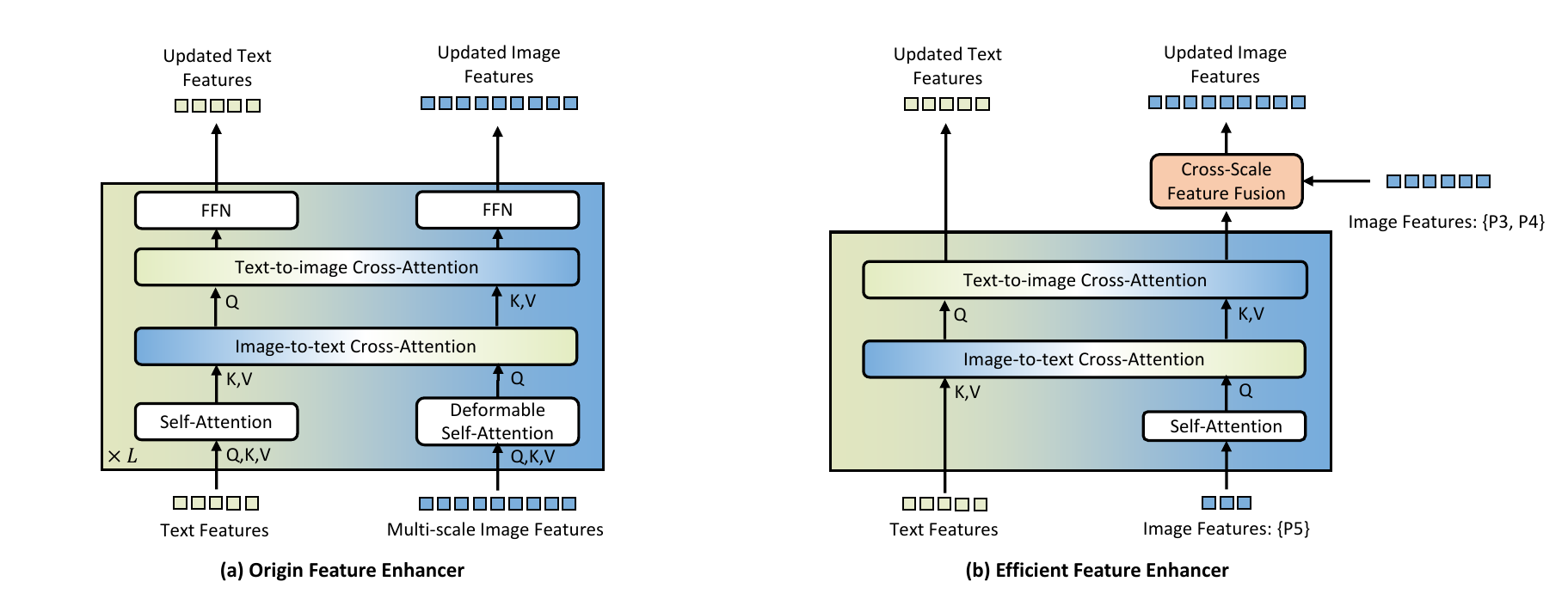}
\centering
\caption{Comparison of the Origin Feature Enhancer and the New Efficient Feature Enhancer.}
\label{fig:efe}
\end{figure}

To overcome this obstacle, we propose a novel efficient feature enhancer, as shown in Fig.\ref{fig:efe}.
Recognizing that lower-level image features lack semantic information and introduce excessive computational costs, as demonstrated in Lite-DETR~\cite{LiteDETR}, we limit cross-modality fusion to high-level image features (P5 level) only. This approach greatly reduces the number of tokens that need to be processed, significantly cutting the computational demands of the feature enhancer. To facilitate easier deployment on edge-side GPUs, we replace deformable self-attention with vanilla self-attention, and introduce a cross-scale feature fusion module~\cite{zhao2023detrs} to integrate low-level image features (from P3 and P4 levels). Such a design effectively balances feature enhancement and computational efficiency.

In our edge-optimized model, Grounding DINO 1.5 Edge, we replace the original feature enhancer with our newly proposed efficient one and employ EfficientViT-L1~\cite{Cai_2023_ICCV} as the image backbone for rapid multi-scale feature extraction. We deploy the model on the NVIDIA Orin NX platform, achieving an inference speed of over 10 FPS at an input size of 640 $\times$ 640. The visualization of the model predictions on NVIDIA Orin NX is shown in Figure~\ref{fig:edge_vis}, which demonstrates the effectiveness of our modifications in real-world edge computing environments.

\subsection{Training Dataset}
To train a robust open-set detector, it is crucial to construct a high-quality grounding dataset that is sufficiently rich in categories and encompasses a wide range of detection scenarios. Grounding DINO 1.5 is pre-trained on over 20M grounding images, termed \textbf{Grounding-20M}, which are collected from publicly available sources. We have carefully developed a series of annotation pipelines and post-processing rules to guarantee the high quality of the grounding annotations.

\section{Model Evaluation}
We compare Grounding DINO 1.5 with other related works on both the zero-shot and fine-tuning settings. The best and the second best results are indicated in \textbf{bold} and with \underline{underline}.

\subsection{Zero-Shot Transfer of Grounding DINO 1.5 Pro}

Following the implementation of Grounding DINO~\cite{GroundingDINO}, we evaluate our model's zero-shot transfer performance on the COCO~\cite{COCO} dataset, which comprises 80 common categories and the LVIS~\cite{LVIS} datasets, which includes 1203 more diverse and long-tailed categories. We report the performance of fixed AP~\cite{FixedAP} on both the LVIS-val and LVIS-minival splits. As shown in Table~\ref{tab:gd-1.5-zero-shot}, our model shows a significant performance improvement compared to the previous Grounding DINO models. For instance, on the COCO zero-shot transfer benchmark, Grounding DINO 1.5 Pro achieves a 54.3 AP, improving upon Grounding DINO Swin-L by 1.8 AP. On the LVIS-minival and LVIS-val zero-shot transfer benchmarks, Grounding DINO 1.5 Pro achieves a 55.7 AP and a 47.6 AP, outperforming the previous best model, DetCLIPv3, by 6.9 AP and 6.2 AP, respectively. Furthermore, compared with the Grounding DINO Swin-T model, our model demonstrates a remarkable improvement of 28.3 AP (55.7 AP vs. 27.4 AP) on the LVIS-minival zero-shot transfer benchmark.

\begin{table}[!h]
\setlength{\tabcolsep}{3pt}
\renewcommand{\arraystretch}{1.25}
\centering
\scriptsize
\resizebox{1.00\linewidth}{!}{
\begin{tabular}{lclccccccccccc}
\toprule
Method & Backbone & Pre-training data & COCO & \multicolumn{4}{c}{LVIS$^{\mathrm{minival}}$} & \multicolumn{4}{c}{LVIS$^{\mathrm{val}}$} & ODinW35 & ODinW13 \\
& & & AP$_{\mathrm{all}}$ &  AP$_{\mathrm{all}}$  & AP$_{\mathrm{r}}$ & AP$_{\mathrm{c}}$ & AP$_{\mathrm{f}}$ & AP$_{\mathrm{all}}$ & AP$_{\mathrm{r}}$ & AP$_{\mathrm{c}}$ & AP$_{\mathrm{f}}$ & AP$_{\mathrm{avg}}$ & AP$_{\mathrm{avg}}$ \\
\midrule
\multicolumn{14}{c}{\textit{Supervised Models (Pre-training data includes COCO, LVIS, etc.)}}\\
\midrule
GLIPv2~\cite{GLIPv2} & Swin-H~\cite{SwinHuge} & FourODs,\textcolor{gray}{COCO},GoldG,CC15M,SBU & \textcolor{gray}{60.6} & \textcolor{gray}{50.1} & - & - & - & - & - & - & - & - & 55.5 \\
Grounding DINO~\cite{GroundingDINO} & Swin-L~\cite{SwinTransformer} & O365,OID,GoldG,Cap4M,\textcolor{gray}{COCO},RefC & \textcolor{gray}{60.7} & \textcolor{gray}{33.9} & \textcolor{gray}{22.2} & \textcolor{gray}{30.7} & \textcolor{gray}{38.8} & - & - & - & - & - & -\\
APE (B)~\cite{APE} & ViT-L & \textcolor{gray}{COCO,LVIS},O365,OID,VG & \textcolor{gray}{57.7} & \textcolor{gray}{62.5} & - & - & - & \textcolor{gray}{57.0} & - & - & - & 29.4 & \sota{59.8} \\
APE (D)~\cite{APE} & ViT-L~\cite{EVA-02} & \textcolor{gray}{COCO,LVIS},O365,OID,VG,RefC,SA-1B,GoldG,PhraseCut & \textcolor{gray}{58.3} & \textcolor{gray}{64.7} & - & - & - & \textcolor{gray}{59.6} & - & - & - & 28.8 & 57.9 \\
GLEE-Pro~\cite{GLEE} & ViT-L~\cite{EVA-02} & GLEE-merged-10M (\textcolor{gray}{COCO,LVIS},etc) & \textcolor{gray}{62.0} & - & - & - & - & \textcolor{gray}{55.7} & \textcolor{gray}{49.2} & - & - & - & 53.4 \\
\midrule
\multicolumn{14}{c}{\textit{Zero-shot Transfer Models}}\\
\midrule
OWL-ViT~\cite{minderer2022simple}  & ViT-L~\cite{vit} & O365,OID,VG,LiT & 42.2  & - & - & - & - & 34.6 & 31.2 & - & - & - & - \\
MDETR~\cite{kamath2021mdetr}  & ResNet101~\cite{ResNet} & COCO,GoldG & -  & 22.5 & 7.4 & 22.7 & 25.0 & - & - & - & - & - 
& - \\
GLIP~\cite{GLIP}  & Swin-L & FourODs,GoldG,Cap24M & 49.8  & 37.3 & 28.2 & 34.3 & 41.5 & 26.9 & 17.1 & 23.3 & 35.4 & - & 52.1 \\
Grounding DINO~\cite{GroundingDINO} & Swin-T & O365,GoldG,Cap4M & 48.4 & 27.4 & 18.1 & 23.3 & 32.7 & - & - & - & - & 22.3 & 49.8 \\
Grounding DINO~\cite{GroundingDINO} & Swin-L & O365,OID,GoldG & {52.5} & - & - & - & - & - & - & - & - & 26.1 & 56.9  \\
OpenSeeD~\cite{zhang2023simple} & Swin-L & COCO,O365 & - & 23.0 & - & - & - & - & - & - & - & 15.2 & -  \\
UniDetector~\cite{wang2023detecting} & ResNet50~\cite{ResNet} & COCO,O365,OID & - & - & - & - & - & 19.8 & 18.0 & 19.2 & 21.2 & - & 47.3  \\
OmDet-Turbo-B~\cite{OmDet_Turbo} & ConvNeXt-B~\cite{ConvNeXt}  &  O365,GoldG,PhraseCut,Hake,HOI-A  & \underline{53.4} & 34.7 & - & - & - & - & - & -  & -  & \underline{30.1} & 54.7 \\
OWL-ST~\cite{OWLViTV2}  & CLIP L/14~\cite{CLIP} & WebLI2B & -  & 40.9 & 41.5 & - & - & 35.2 & 36.2 & - & - & 24.4 & 53.0 \\
MQ-GLIP~\cite{xu2024multi}  & Swin-L & O365 & -  & 43.4 & 34.5 & 41.2 & 46.9 & 34.7 & 26.9 & 32.0 & 41.3 & 23.9 & 54.1 \\
DetCLIP~\cite{DetCLIP}  & Swin-L & O365,GoldG,YFCC1M & -  & 38.6 & 36.0 & 38.3 & 39.3 & 28.4 & 25.0 & 27.0 & 31.6 & - & - \\
DetCLIPv2~\cite{DetCLIPv2}  & Swin-L & O365,GoldG,CC15M & -  & 44.7 & 43.1 & 46.3 & 43.7 & 36.6 & 33.3 & 36.2 & 38.5 & - & - \\
DetCLIPv3~\cite{DetCLIPv3} & Swin-L & O365,V3Det,GoldG,GranuCap50M & -   & 48.8 & \underline{49.9} & 49.7 & 47.8 & 41.4 & 41.4 & 40.5 & 42.3 & - & - \\
YOLO-World~\cite{cheng2024yolo} & YOLOv8-L~\cite{YOLOv8} & O365,GoldG,CC3M & 45.1 & 35.4 & 27.6 & 34.1 & 38.0 & - & - & - & - & - & - \\
DINOv ~\cite{li2023visual} & Swin-L &COCO,SA-1B & - & - & - & - & - & - & - & - & - & 15.7 & - \\
T-Rex2 (visual) ~\cite{TRex2} & Swin-L &O365,OID,HierText,CrowdHuman,SA-1B & 46.5 & 47.6 & 45.4 & 46.0 & 49.5 & 45.3 & \underline{43.8} & 42.0 & 49.5 & 27.8 & - \\
T-Rex2 (text) ~\cite{TRex2} & Swin-L &O365,OID,GoldG,CC3M,SBU,LAION & 52.2 & \underline{54.9} & 49.2 & \underline{54.8} & \sota{56.1} & \underline{45.8} & 42.7 & \underline{43.2} & \sota{50.2} & 22.0 & - \\
\midrule
Grounding DINO 1.5 Pro \tiny{(zero-shot)} & ViT-L~\cite{EVA-02} & Grounding-20M  & \sota{54.3}  & \sota{55.7} & \sota{56.1} & \sota{57.5} & \underline{54.1} & \sota{47.6} & \sota{44.6} & \sota{47.9} & \underline{48.7} & \sota{30.2} & \underline{58.7} \\
\bottomrule
\end{tabular}
}
\caption{Performance of Grounding DINO 1.5 Pro on the COCO, LVIS, ODinW35~\cite{GLIP} and ODinW13~\cite{GLIP} benchmarks compared to previous methods. \textcolor{gray}{Gray} numbers indicate that the training dataset includes images or annotations from COCO or LVIS datasets.}
\label{tab:gd-1.5-zero-shot}
\end{table} 

\begin{table}[!h]
\setlength{\tabcolsep}{3pt}
\renewcommand{\arraystretch}{1.25}
\centering
\scriptsize
\resizebox{1.0\linewidth}{!}{
\begin{tabular}{lccccccccccccccc}
\toprule
Method & Backbone & PascalVOC & AerialDrone & Aquarium & Rabbits & EgoHands & Mushrooms & Packages & Raccoon & Shellfish & Vehicles & Pistols & Pothole & Thermal & AP$_{\mathrm{avg}}$ \\
\midrule
GLIP & Swin-L & 61.7 & 7.1 & 26.9 & 75.0 & 45.5 & 49.0 & 62.8 & 63.3 & \sota{68.9} & 57.3 & 68.6 & 25.7 & 66.0 & 52.1 \\
GLIPv2 & Swin-H & 66.3 & 10.9 & 30.4 & 74.6 & 55.1 & 52.1 & 71.3 & 63.8 & \underline{66.2} & 57.2 & 66.4 & \sota{33.8} & 73.3 & 55.5 \\
Grounding DINO & Swin-L & 66.0 & 12.6 & 28.1 & 72.8 & 52.1 & 73.0 & 63.9 & \underline{67.9} & 64.8 & 62.7 & 71.7 & \underline{31.4} & \sota{78.4} & 56.9 \\
OmDet-Turbo-B & ConvNeXt-B & 63.7 & 16.2 & 28.5 & 70.6 & 55.7 & 71.5 & \underline{65.6} & 63.6 & 39.7 & 61.9 & 65.5 & 30.2 & \sota{78.4} & 54.7 \\
OWL-ST & CLIP L/14 & 53.9 & 19.9 & 32.3 & \sota{84.9} & 47.1 & 76.6 & \sota{70.9} & 63.8 & 35.0 & 58.5 & 62.6 & 27.5 & 55.6 & 53.0 \\
MQ-GLIP-L & Swin-L & 64.7 & 17.4 & 30.3 & 71.8 & 57.2 & 63.9 & 53.0 & 58.1 & 63.0 & 63.2 & \sota{74.4} & 27.0 & 58.7 & 54.1 \\
APE (B) & ViT-L & - & - & - & - & - & - & - & - & - & - & - & - & - & \sota{59.8} \\
APE (D) & ViT-L & - & - & - & - & - & - & - & - & - & - & - & - & - & 57.9 \\
GLEE-Pro-Scale & ViT-L & 69.1 & 13.7 & 34.7 & \underline{75.6} & 38.9 & 57.8 & 50.6 & 65.6 & 62.7 & \sota{67.3} & 69.0 & 30.7 & 59.1 & 53.4 \\
T-Rex2 (visual prompt) & Swin-L & 65.8 & 16.0 & 27.0 & 70.0 & \sota{61.9} & \sota{83.7} & 58.9 & 67.1 & 53.0 & \underline{66.4} & 69.0 & 24.1 & 61.4 & 55.7 \\
\midrule
Grounding DINO 1.5 Pro & ViT-L & \underline{67.1} & \underline{19.0} & \sota{38.5} & 65.7 & \underline{61.8} &  \underline{82.1} & 58.1 & \sota{72.5} & 62.0 & {64.3} & \underline{71.9} & 29.0 & \underline{71.4} & \underline{58.7} \\
\bottomrule
\end{tabular}
}
\caption{Detailed zero-shot performance comparison between Grounding DINO 1.5 Pro and related works on ODinW13 benchmark.}
\label{tab:odinw13_zero_shot}
\end{table} 

We further evaluate the generalization capability of our model in multiple real-world scenarios using the ODinW (Object Detection in the Wild)~\cite{GLIP} benchmark, which encompasses 35 datasets covering a wide range of application domains. We observe that within the ODinW benchmark, several datasets exhibit significant quality issues in terms of the annotated category names. To mitigate such problems, during testing, we performed prompt engineering to refine category names on datasets where their performance was particularly poor to better align their category names with actual testing scenarios. Grounding DINO 1.5 Pro achieves an average of 58.7 AP over 13 datasets on the ODinW13 benchmark and set a new record on the ODinW35 benchmark with an average of 30.2 AP over 35 datasets, improving upon Grounding DINO by 4.2 AP. The comprehensive per-dataset performance of Grounding DINO 1.5 Pro on ODinW13 are presented in Table~\ref{tab:odinw13_zero_shot}. Furthermore, the detailed per-dataset performance of Grounding DINO 1.5 Pro on ODinW35 is available in Appendix Section~\ref{odinw_detailed_performance}.

\subsection{Fine-tuning Results on Downstream Datasets}

We further investigate the transferability of Grounding DINO 1.5 Pro by fine-tuning it on various downstream datasets. As shown in Table~\ref{tab:downstream_finetune}, on the LVIS dataset, the fine-tuned Grounding DINO 1.5 Pro model achieves a 68.1 AP and a 63.5 AP on the LVIS-minival and LVIS-val splits, respectively, which represent an enhancement of 12.4 AP and 15.9 AP over the Grounding DINO 1.5 Pro zero-shot setting.

\begin{table}[!h]
\setlength{\tabcolsep}{3pt}
\renewcommand{\arraystretch}{1.25}
\centering
\scriptsize
\resizebox{1.0\linewidth}{!}{
\begin{tabular}{lccccccccccc}
\toprule
Method & Backbone & \multicolumn{4}{c}{LVIS$^{\mathrm{minival}}$} & \multicolumn{4}{c}{LVIS$^{\mathrm{val}}$} & ODinW35 & ODinW13 \\
&  &  AP$_{\mathrm{all}}$  & AP$_{\mathrm{r}}$ & AP$_{\mathrm{c}}$ & AP$_{\mathrm{f}}$ & AP$_{\mathrm{all}}$ & AP$_{\mathrm{r}}$ & AP$_{\mathrm{c}}$ & AP$_{\mathrm{f}}$ & AP$_{\mathrm{avg}}$ & AP$_{\mathrm{avg}}$ \\
\midrule
GLIP & Swin-L & - & - & - & - & - & - & - & - & - & 68.9 \\
GLEE-Pro & ViT-L & - & - & - & - & - & - & - & - & - & 69.0 \\
GLIPv2 & Swin-H  & 59.8 & - & - & - & - & - & - & - & - & 70.4 \\
OWL-ST+FT \dag & CLIP L/14  & 54.4 & 46.1 & - & - & 49.4 & 44.6 & - & - & - & - \\
DetCLIPv2 & Swin-L & 60.1 & 58.3 & 61.7 & 59.1 & 53.1 & \underline{49.0} & 53.2 & 54.9 & - & 70.4 \\
DetCLIPv3 & Swin-L & 60.5 & \underline{60.7} & - & - & - & - & - & - & - & \underline{72.1} \\
DetCLIPv3 \dag & Swin-L & \underline{60.8} & 56.7 & \underline{63.2} & \underline{59.4} & \underline{54.1} & 45.8 & \underline{55.4} & \underline{56.4} & - & - \\
\midrule
Grounding DINO 1.5 Pro \tiny{(zero-shot)} & ViT-L & {55.7} & {56.1} & {57.5} & {54.1} & {47.6} & {44.6} & {47.9} & {48.7} & 30.2 & 58.7 \\
Grounding DINO 1.5 Pro & ViT-L & \sota{68.1} \textbf{\textcolor{red}{(+12.4)}} & \sota{68.7} & \sota{69.5} & \sota{66.6} & \sota{63.5} \textbf{\textcolor{red}{(+15.9)}} & \sota{64.0} & \sota{63.8} & \sota{63.0} & \sota{70.6} \textbf{\textcolor{red}{(+40.4)}} & \sota{72.4} \textbf{\textcolor{red}{(+13.7)}} \\
\bottomrule
\end{tabular}
}
\caption{Fine-tuning performance of Grounding DINO 1.5 Pro on the LVIS-minival, LVIS-val, ODinW35 and ODinW13 benchmarks. The fixed AP~\cite{FixedAP} on LVIS-minival and val splits are reported. \dag indicates results of fine-tuning with LVIS base categories only.}
\label{tab:downstream_finetune}
\end{table} 

After fine-tuning on the ODinW35 dataset, the Grounding DINO 1.5 Pro model sets new records by achieving an average of 70.6 AP across 35 datasets on the ODinW35 benchmark and 72.4 AP over 13 datasets on the ODinW13 benchmark, respectively. This represents a significant improvement over the zero-shot setting, with respective improvements of 40.5 AP and 13.7 AP. The detailed fine-tuning performance of Grounding DINO 1.5 Pro on the ODinW13 benchmark is reported in Table~\ref{tab:downstream_finetune_odinw13}. 

\begin{table}[!h]
\setlength{\tabcolsep}{3pt}
\renewcommand{\arraystretch}{1.25}
\centering
\scriptsize
\resizebox{1.0\linewidth}{!}{
\begin{tabular}{lccccccccccccccc}
\toprule
Method & Backbone & PascalVOC & AerialDrone & Aquarium & Rabbits & EgoHands & Mushrooms & Packages & Raccoon & Shellfish & Vehicles & Pistols & Pothole & Thermal & AP$_{\mathrm{all}}$ \\
\midrule
GLIP & Swin-L & 69.6 & 32.6 & 56.6 & 76.4 & 79.4 & 88.1 & 67.1 & 69.4 & 65.8 & 71.6 & 75.7 & 60.3 & 83.1 & 68.9 \\
GLEE-Pro & ViT-L & 72.6 & 36.5 & 58.1 & \underline{80.5} & 74.1 & \underline{92.0} & 67.0 & 76.5 & 66.4 & 70.5 & 66.4 & 55.7 & 80.6 & 69.0 \\
GLIPv2 & Swin-H & 74.4 & 36.3 & \underline{58.7} & 77.1 & 79.3 & 88.1 & 74.3 & 73.1 & \underline{70.0} & \underline{72.2} & 72.5 & 58.3 & 81.4 & 70.4 \\
DetCLIPv2 & Swin-L & 74.4 & \underline{44.1} & 54.7 & \sota{80.9} & \underline{79.9} & 90.0 & 74.1 & 69.4 & 61.2 & 68.1 & \underline{80.3} & 57.1 & 81.1 & 70.4 \\
Grounding DINO & Swin-T & 73.6 & 36.6 & 57.7 & 78.7 & 79.2 & \sota{92.8} & \underline{74.7} & 74.7 & 61.2 & 69.6 & 75.9 & \underline{60.4} & \sota{85.9} & 70.9 \\
MQ-GLIP-L & Swin-L & - & - & - & - & - & - & - & - & - & - & - & - & - & 71.3 \\
DetCLIPv3 & Swin-L & \underline{76.4} & \sota{51.2} & 57.5 & {79.9} & \sota{80.2} & {90.4} & \sota{75.1} & 70.9 & 63.6 & 69.8 & \sota{82.7} & 56.2 & 83.8 & \underline{72.1} \\
\midrule
Grounding DINO 1.5 Pro & ViT-L & \sota{77.6} & 37.0 & \sota{60.2} & 75.1 & 78.6 & 89.2 & 72.1 & \sota{81.8} & \sota{70.8} & \sota{74.6} & 77.6 & \sota{62.3} & \underline{84.0} & \sota{72.4} \\
\bottomrule
\end{tabular}}
\caption{Detailed fine-tuning performance of Grounding DINO 1.5 Pro on the ODinW13 benchmark.}
\label{tab:downstream_finetune_odinw13}
\end{table} 

\subsection{Main Results of Grounding DINO 1.5 Edge}
After pre-training on Grounding-20M, we directly evaluate Grounding DINO 1.5 Edge on the COCO dataset and LVIS dataset in a zero-shot manner. Following previous works\cite{GroundingDINO,cheng2024yolo,GLIPv2}, we evaluate on both LVIS-val and LVIS-minival splits and report the fixed AP~\cite{FixedAP} for comparison. The main results are shown in Table~\ref{tab:edge}. 
Compared with current real-time open-set detectors in an end-to-end test setting, which do not use language cache, Grounding DINO 1.5 Edge achieves a zero-shot AP of 45.0 on COCO.  Regarding the zero-shot performance on LVIS-minival, Grounding DINO 1.5 Edge achieves remarkable performance, an AP score of 36.2, which surpasses all other state-of-the-art algorithms (OmDet-Turbo-T 30.3 AP, YOLO-Worldv2-L 32.9 AP, YOLO-Worldv2-M 30.0 AP, YOLO-Worldv2-S 22.7 AP).  Notably, deploying Grounding DINO 1.5 Edge model optimized with TensorRT on NVIDIA Orin NX achieves an inference speed of over 10 FPS at an input size of 640 $\times$ 640.
\begin{table}[!h]
\setlength{\tabcolsep}{3pt}
\renewcommand{\arraystretch}{1.25}
\centering
\scriptsize
\resizebox{1.0\linewidth}{!}{
\begin{tabular}{lclcccccccccccc}
\toprule
Method & Backbone & Pre-training data & test size & COCO & \multicolumn{4}{c}{LVIS$^{\mathrm{minival}}$} & \multicolumn{4}{c}{LVIS$^{\mathrm{val}}$} & FPS(A100/TensorRT) & FPS(Orin NX) \\
& & & & & AP$_\mathrm{all}$  & $AP_\mathrm{r}$ & $AP_\mathrm{c}$ & $AP_\mathrm{f}$ & $AP_\mathrm{all}$ & $AP_\mathrm{r}$ & $AP_\mathrm{c}$ & $AP_\mathrm{f}$ &  &\\
\midrule
\multicolumn{14}{c}{\textit{End-to-End Open-Set Object Detection}}\\
\midrule
GLIP-T & Swin-T & O365,GoldG,Cap4M & 800 $\times$ 1333 & 46.3 & 26.0 & 20.8 & 21.4 & 31.0 & - & - & - & - & - & - \\
Grounding DINO-T & Swin-T & O365,GoldG,Cap4M & 800 $\times$ 1333 & 48.4 & 27.4 & 18.1 & 23.3 & 32.7 & - & - & - & - &  9.4 / 42.6 & 1.1 \\ 
\midrule
\multicolumn{14}{c}{\textit{Real-time End-to-End Open-Set Object Detection}}\\
\midrule
YOLO-Worldv2-S\dag  & YOLOv8-S & O365,GoldG & 640 $\times$ 640 & - & 22.7 & 16.3 & 20.8 & 25.5 & 17.3 & 11.3 & 14.9	& 22.7 & 47.4 / - & - \\
YOLO-Worldv2-M\dag  & YOLOv8-M & O365,GoldG & 640 $\times$ 640 & - & 30.0 & 25.0 & 27.2 & 33.4 & 23.5 & 17.1 & 20.0	& 30.1 & 42.7 / - & - \\
YOLO-Worldv2-L\dag  & YOLOv8-L & O365,GoldG & 640 $\times$ 640 & - & 33.0	& 22.6 & 32.0 & 35.8 & 26.0 & 18.6 & 23.0 & \textbf{32.6} & 37.4 / - & - \\
YOLO-Worldv2-L\dag  & YOLOv8-L & O365,GoldG,CC3M-Lite & 640 $\times$ 640 & - & 32.9	& 25.3 & 31.1 & \underline{35.8} & 26.1 & 20.6 & 22.6 & \underline{32.3} & 37.4 / - & - \\
OmDet-Turbo-T$^\ddagger$  & Swin-T  & O365,GoldG & 640 $\times$ 640 & 42.5 & 30.3 & - & - & - & - & - & - & - & 21.5 / 140.0 & - \\
\midrule
Grounding DINO 1.5 Edge & EfficientViT-L1 & Grounding-20M & 640 $\times$ 640 & \underline{42.9} & \underline{33.5} & \underline{28.0} & \underline{34.3} & 33.9 & \underline{27.3} & \underline{26.3} & \underline{25.7} & 29.6 & 21.7 / 111.6 & 10.7 \\ 
Grounding DINO 1.5 Edge & EfficientViT-L1 & Grounding-20M &  800 $\times$ 1333 & \textbf{45.0} & \textbf{36.2} & \textbf{33.2} & \textbf{36.6} & \textbf{36.3} & \textbf{29.3} & \textbf{28.1} & \textbf{27.6} & 31.6 & 18.5 / 75.2 & 5.5 \\ 
\bottomrule
\end{tabular}}
\caption{Zero-shot Results of Grounding DINO 1.5 Edge on COCO and LVIS. Speed measurement is performed on an A100 GPU, expressed in frames per second (FPS). The format used is PyTorch speed / TensorRT FP32 speed. And FPS on NVIDIA Orin NX is also reported. \dag indicates results of YOLO-World are reproduced by the latest official codes. $^\ddagger$ indicates it uses language cache, which does not calculate the latency of the text encoder.}
\label{tab:edge}
\end{table}

\vspace{-6mm}
\section{Case Analysis and Qualitative Visualization}

In this section, we visualize the detection results of Grounding DINO 1.5 models in real-world scenarios. The images and text prompts are primarily sourced from the COCO~\cite{COCO}, LVIS~\cite{LVIS}, V3Det~\cite{V3Det}, OpenImages~\cite{OpenImages}, CC3M~\cite{CC12M} and SA-1B~\cite{SAM}. We are deeply grateful for their contributions, which have significantly benefited the community.

\subsection{Common Object Detection}
\begin{figure}[ht!]
\centering
 \includegraphics[width=1.0\textwidth,keepaspectratio]{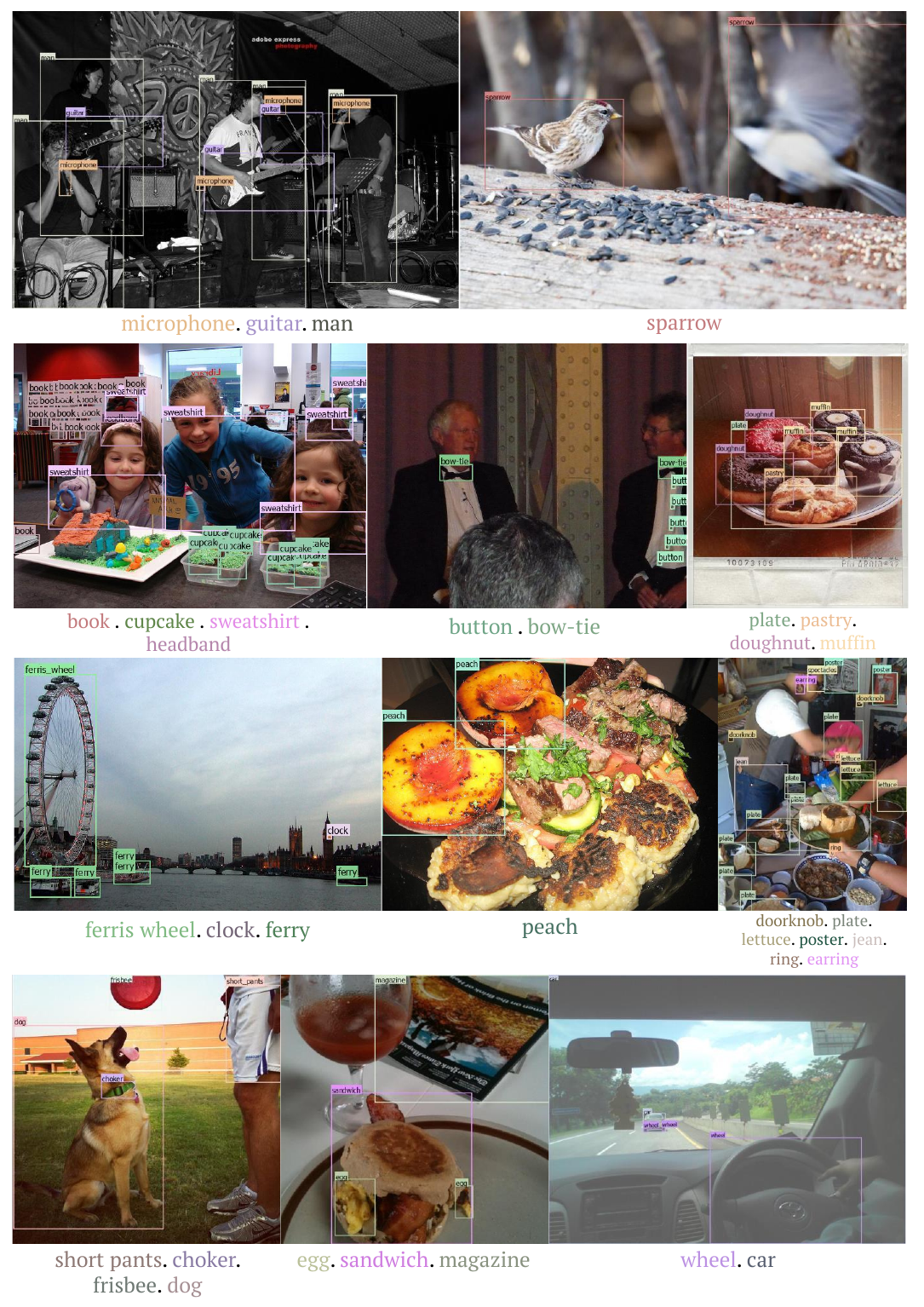}
 \caption{Model predictions on common objects with Grounding DINO 1.5 Pro (part 1).}
 \label{fig:vis_common_obj}
\end{figure}

\begin{figure}[ht!]
\centering
 \includegraphics[width=1.0\textwidth,keepaspectratio]{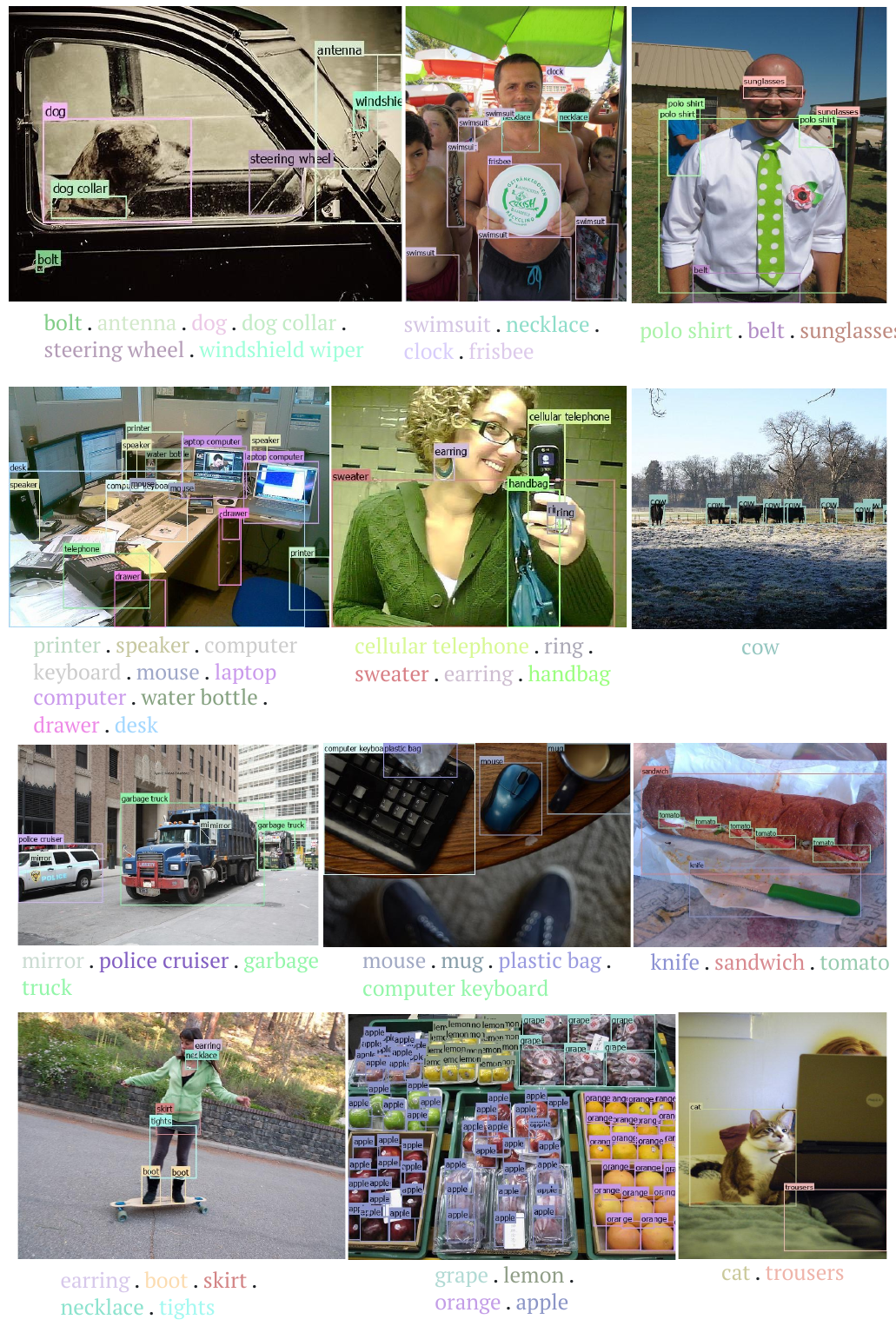}
 \caption{Model predictions on common objects with Grounding DINO 1.5 Pro (part 2).}
 \label{fig:vis_common_obj2}
\end{figure}

The visualizations presented in Figures \ref{fig:vis_common_obj} and \ref{fig:vis_common_obj2} demonstrate the robust performance of Grounding DINO 1.5 Pro for common object detection scenarios. Our model's proficiency is evident not only in its handling of typical cases but also in its ability to accurately detect objects under challenging conditions.

The model adeptly identifies objects in monochromatic images, where color cues are minimal, as illustrated by the first example in Figure \ref{fig:vis_common_obj}. This showcases the model's reliance on shape and texture to distinguish objects. The detection of blurry objects, as seen in the second example of Figure \ref{fig:vis_common_obj}, is a testament to the model's robustness against common image degradations, maintaining high accuracy even when visual clarity is compromised.

The ability to detect small and partially occluded objects is crucial for many applications. Grounding DINO 1.5 Pro's success in these scenarios, as shown in the last image in Figure \ref{fig:vis_common_obj}, indicates its fine-grained understanding and the nuanced integration of multi-modal information in autonomous driving scenes. The visualizations also highlight the model's versatility in handling objects of varying sizes and shapes, from petite to sprawling, each accurately localized and identified.

\subsection{Long-tailed Object Detection}

\begin{figure}[ht!]
\centering
 \includegraphics[width=\textwidth,keepaspectratio]{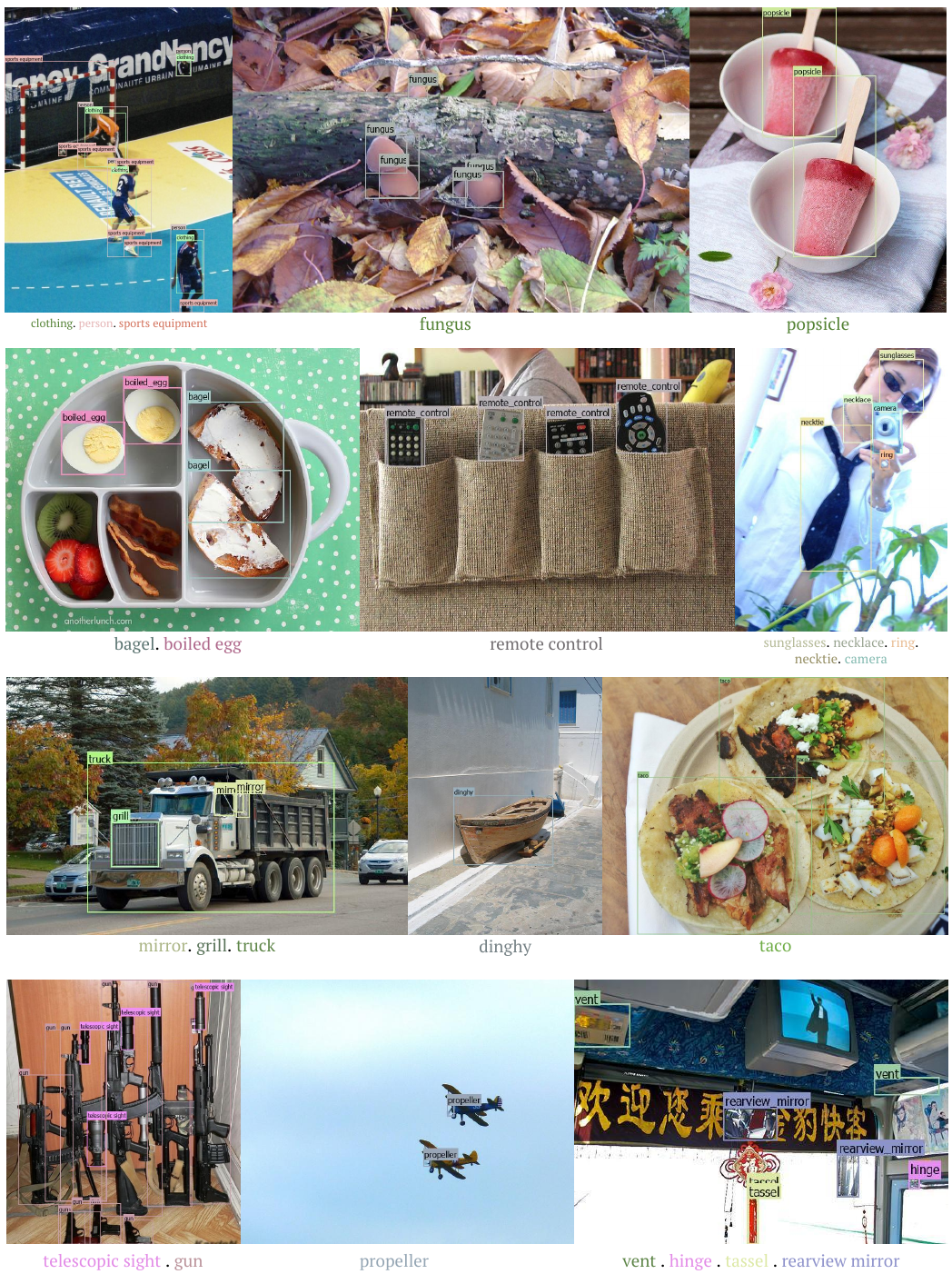}
 \caption{Model predictions on long-tailed categories with Grounding DINO 1.5 Pro.
}
 \label{fig:vis_long_tail}
\end{figure}

This subsection delves into the capability of Grounding DINO 1.5 Pro in detecting long-tailed objects, which are less frequently encountered categories that pose unique challenges for object detection models. The examples provided in Figure \ref{fig:vis_long_tail} highlight the model's nuanced capability of understanding and detecting such uncommon objects. The model demonstrates its ability to recognize a diverse set of objects, including those that are not commonly found in everyday settings. 

For instance, the second image within the figure illustrates the model's capacity to identify a fungus, which is a category that requires specialized knowledge to discern. The third image showcases the model's fine-grained detection capabilities, accurately pinpointing a popsicle amidst a variety of potential distractors. The model's contextual understanding is evident in its ability to detect a taco in the last image of the third line, an object that may be challenging to recognize without the appropriate contextual cues.

\subsection{Short Caption Grounding}

\begin{figure}[ht!]
\centering
 \includegraphics[width=0.92\textwidth,keepaspectratio]{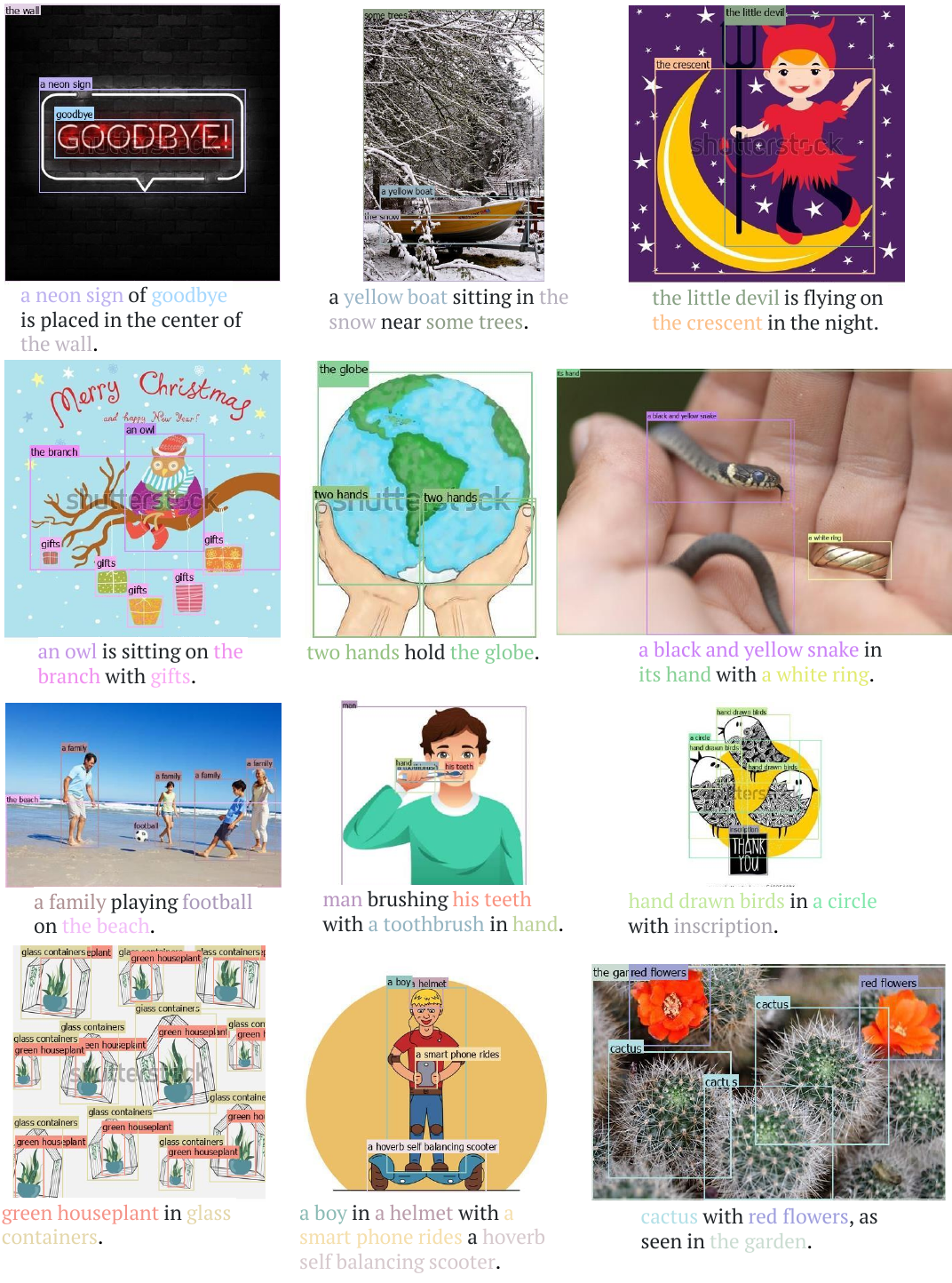}
 \caption{Phrase grounding on short captions with Grounding DINO 1.5 Pro.
}
 \label{fig:vis_short_cap}
\end{figure}

Grounding models can correlate objects within images to their corresponding mentions in accompanying captions. This capability is particularly significant for enhancing the contextual understanding of visual content across various domains. In Figure \ref{fig:vis_short_cap}, we present Grounding DINO 1.5 Pro's proficiency in short caption grounding, highlighting its versatility and accuracy.

The model exhibits a robust performance in grounding objects across different visual domains. It adeptly handles real-world images while also demonstrating a keen ability to interpret objects within cartoons, sketches, and animations. By aligning textual descriptions with visual features, the model can accurately identify and localize objects, even when they appear in stylized or abstract forms.

\subsection{Long Caption Grounding}

\begin{figure}[ht!]
\centering
\includegraphics[width=1.0\textwidth,keepaspectratio]{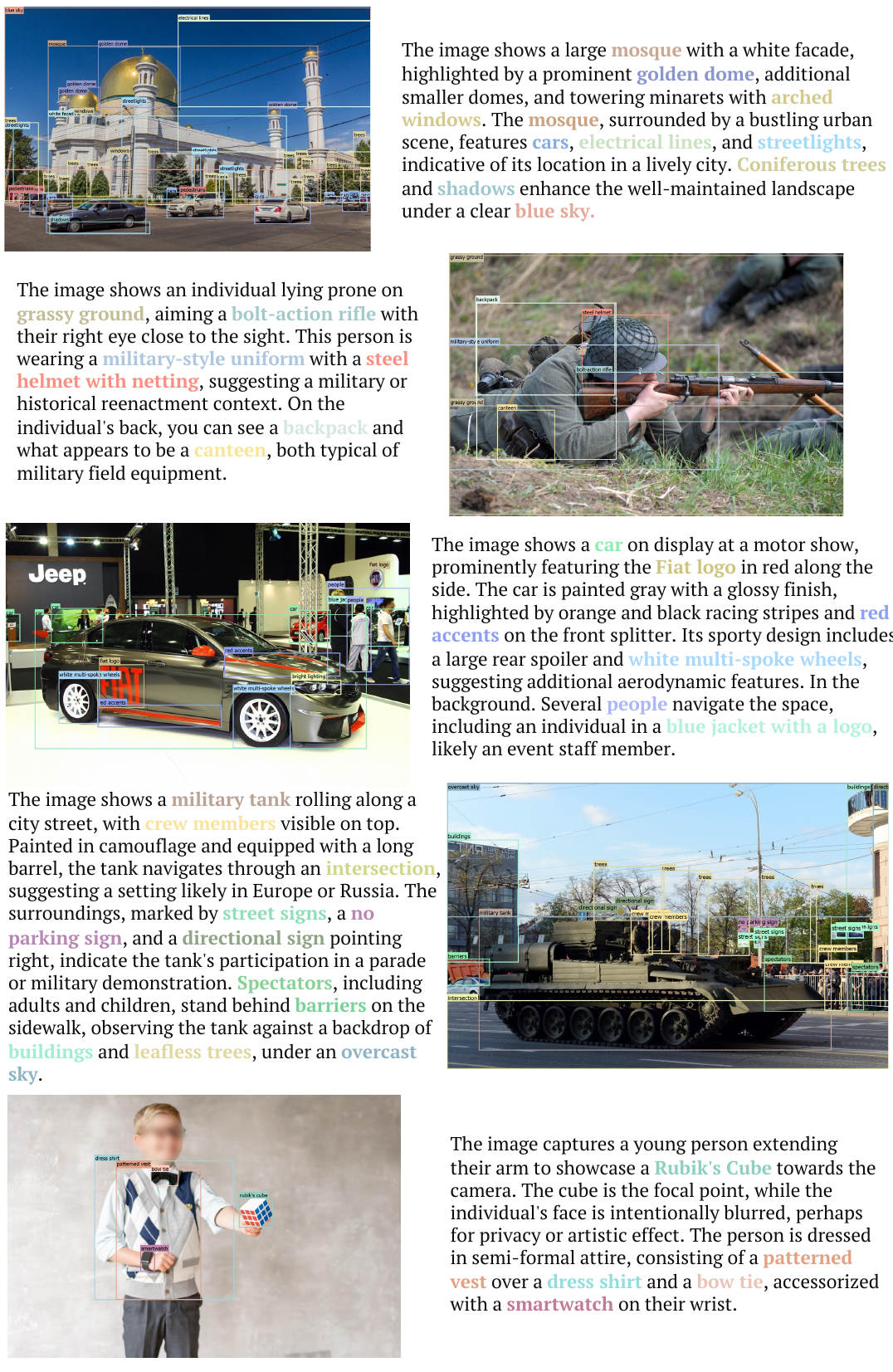}
 \caption{Phrase grounding on long captions with Grounding DINO 1.5 Pro (part 1).
}
 \label{fig:vis_long_cap1}
\end{figure}

\begin{figure}[ht!]
\centering
 \includegraphics[width=0.98\textwidth,keepaspectratio]{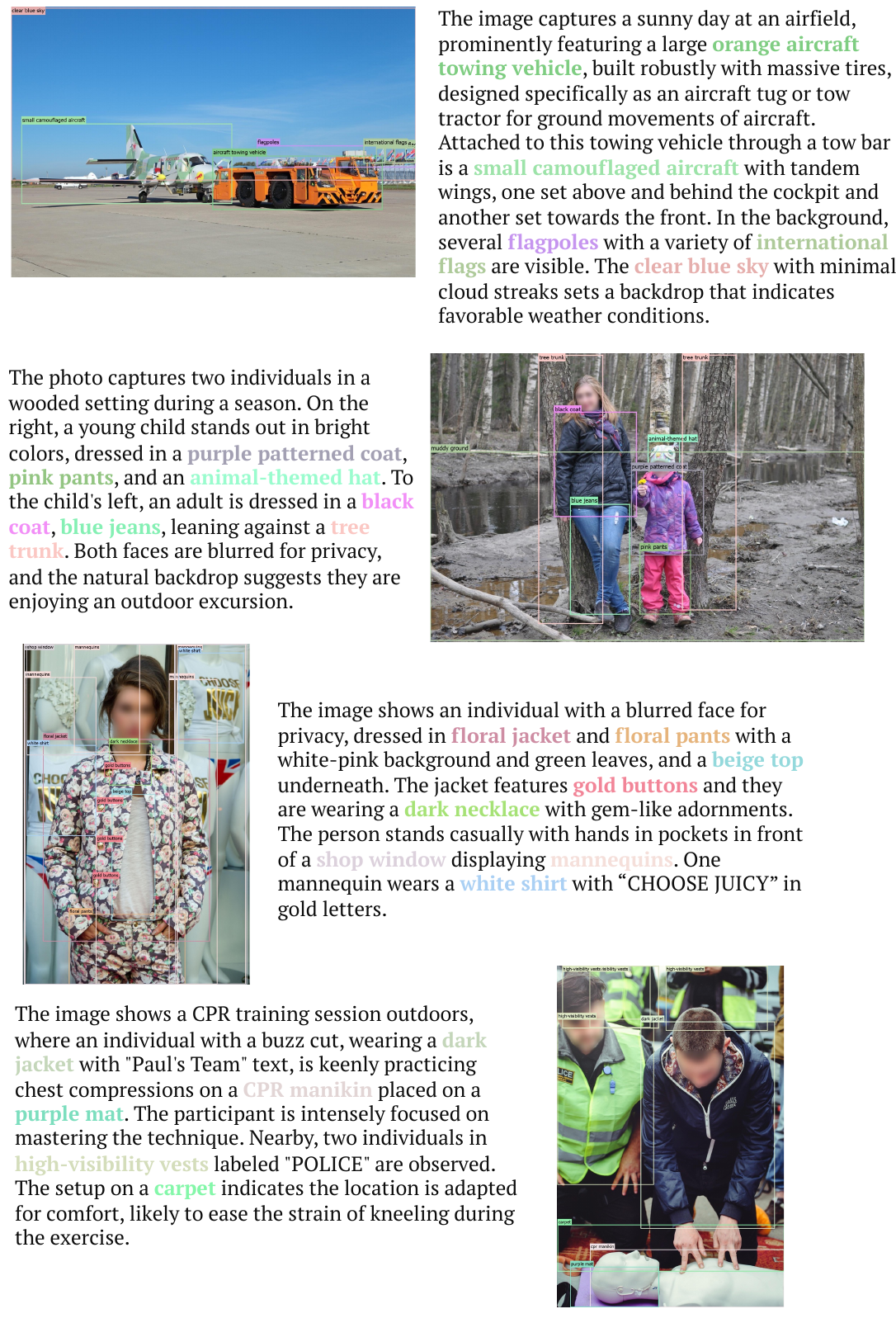}
 \caption{Phrase grounding on long captions with Grounding DINO 1.5 Pro (part 2). 
}
 \label{fig:vis_long_cap2}
\end{figure}

\begin{figure}[ht!]
\centering
 \includegraphics[width=0.98\textwidth,keepaspectratio]{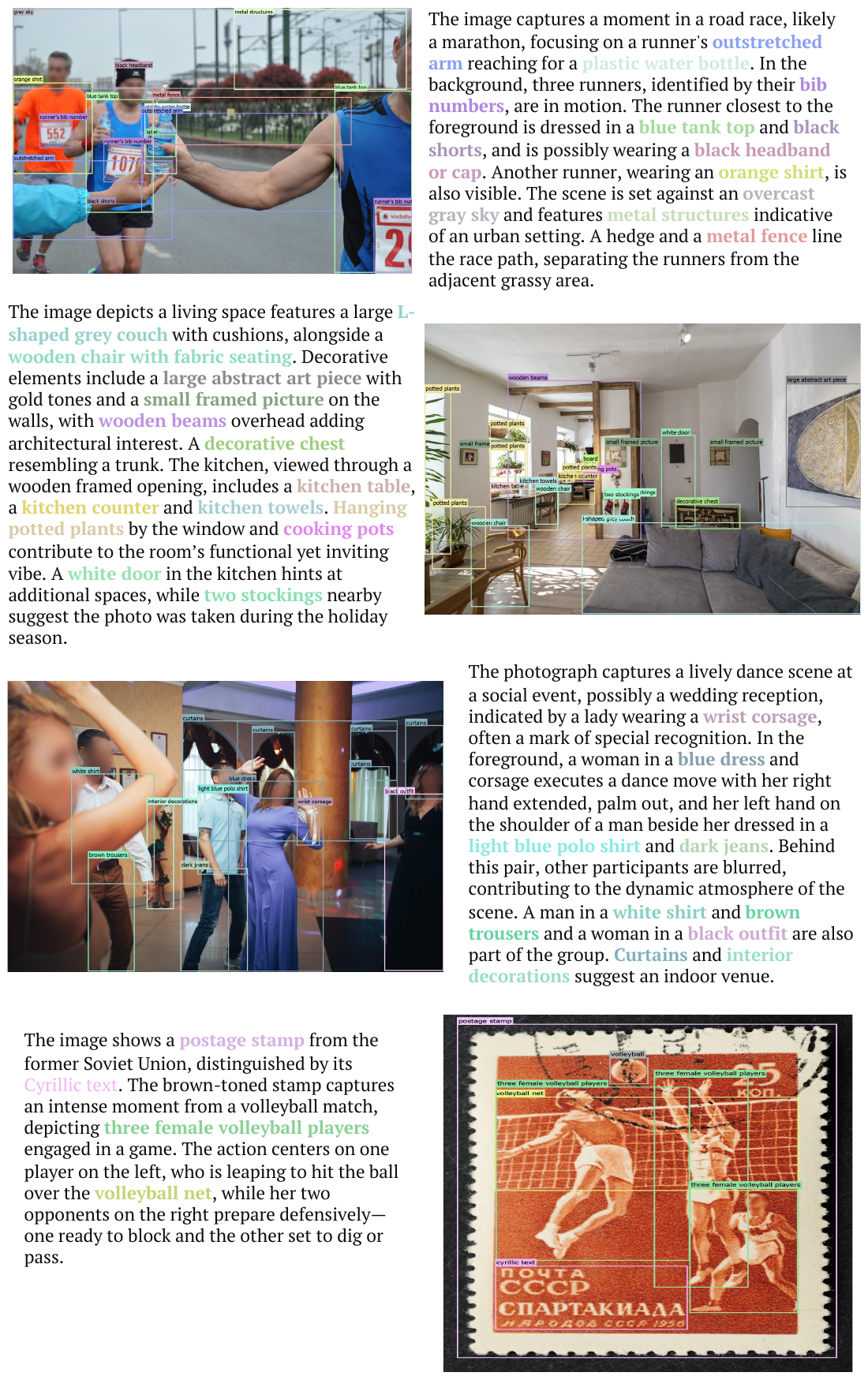}
 \caption{Phrase grounding on long captions with Grounding DINO 1.5 Pro (part 3).
}
 \label{fig:vis_long_cap3}
\end{figure}

Our model, Grounding DINO 1.5 Pro, extends its capability beyond the realm of standard image-caption pairs to adeptly handle long captions, as depicted in Figure \ref{fig:vis_long_cap1}, Figure \ref{fig:vis_long_cap2} and Figure \ref{fig:vis_long_cap3}. Long captions offer a richer tapestry of details that can more comprehensively describe the visual content of an image. The ability to map each noun phrase in a long caption to corresponding objects within an image is a significant step toward deeper image understanding.

An intriguing observation is the model's ability to generalize from pre-training on captions with shorter context windows to effectively processing longer contexts. This adaptability suggests that larger models may inherently possess flexibility that can be leveraged for various lengths of textual input.

Grounding DINO 1.5 Pro exhibits an impressive capacity to correlate textual phrases with visual elements. This capability is not only showcased in the model's handling of long captions but also in its nuanced analysis of images at multiple granularities. Moreover, we notice that the model can detect some terms that did not show in the training data, like \texttt{fiat logo} in the third image. It presents the strong generalization ability of the model.

\subsection{Dense Object Detection}

\begin{figure}[ht!]
 \centering
 \includegraphics[width=1.0\textwidth,keepaspectratio]{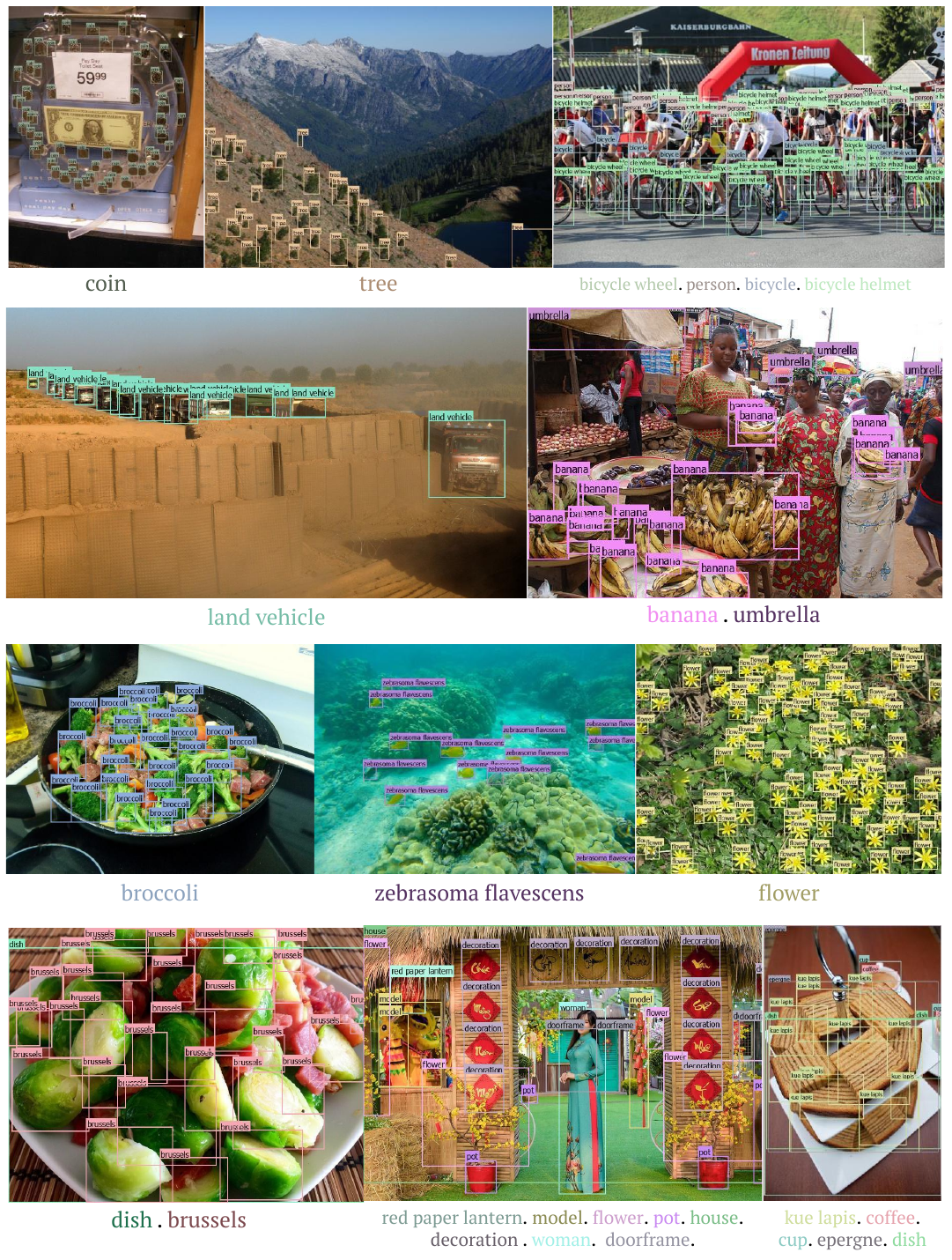}
 \caption{Model predictions on dense object scenarios with Grounding DINO 1.5 Pro (part 1).
}
 \label{fig:vis_dense_obj}
\end{figure}

\begin{figure}[ht!]
\centering
 \includegraphics[width=\textwidth,keepaspectratio]{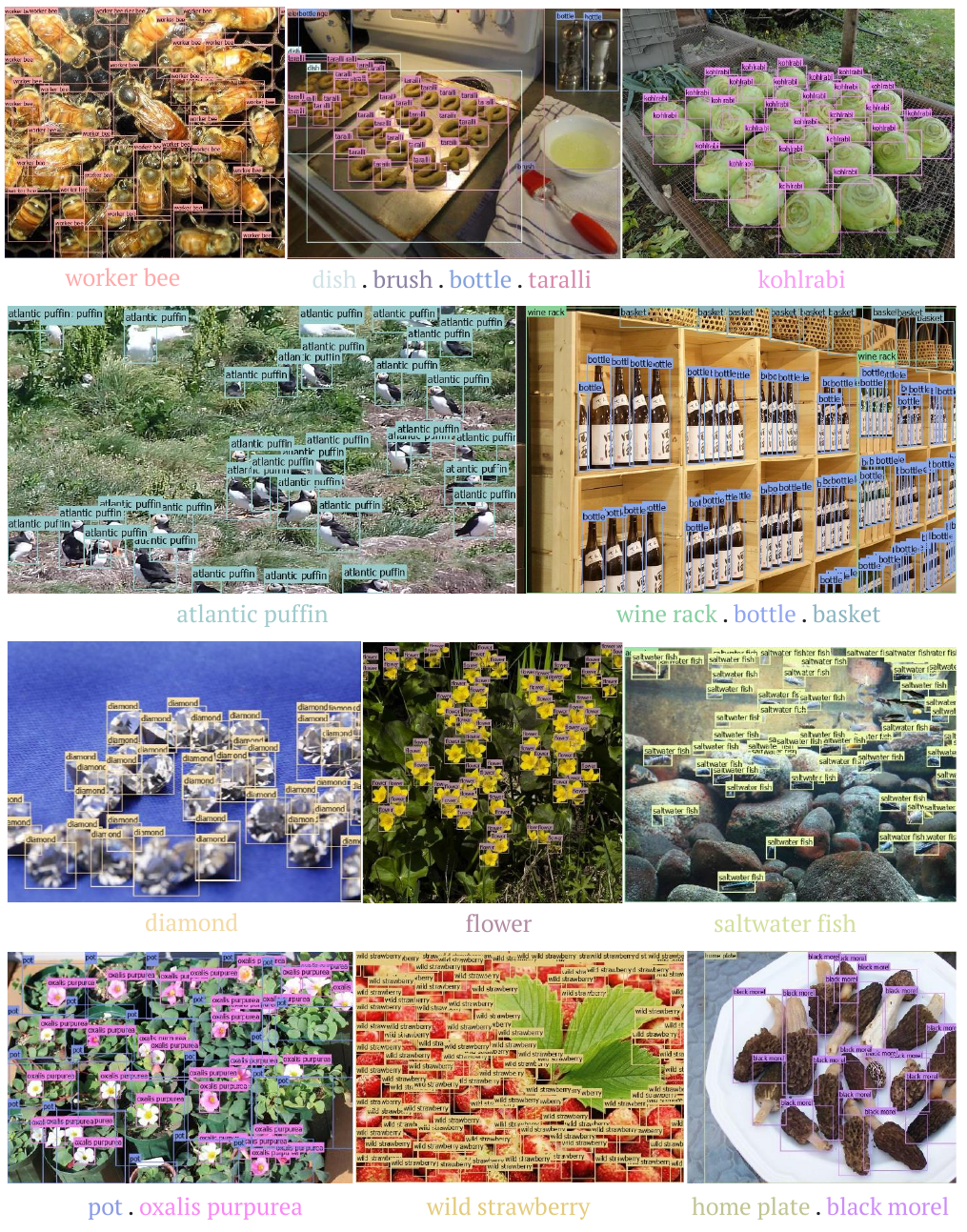}
 \caption{Model predictions on dense object scenarios with Grounding DINO 1.5 Pro (part 2).
}
 \label{fig:vis_dense_obj2}
\end{figure}

Grounding DINO 1.5 Pro showcases an exceptional capability to discern objects within dense scenarios, where multiple objects are closely positioned or overlapping, making detection a challenging task. This ability is vividly demonstrated through the visualizations presented in Figure \ref{fig:vis_dense_obj} and Figure \ref{fig:vis_dense_obj2}.

The model's performance is noteworthy across a wide spectrum of object nomenclature. It adeptly identifies objects labeled with common names such as \texttt{coin}, \texttt{tree}, \texttt{flower}, and \texttt{land}. Moreover, the model also excels at recognizing objects denoted by specialized terminology, including \texttt{kohlrabi}, \texttt{atlantic puffin}, and \texttt{oxalis purpurea}.

\subsection{Video Object Detection}

\begin{figure}[ht!]
\centering
 \includegraphics[width=1.0\textwidth,keepaspectratio]{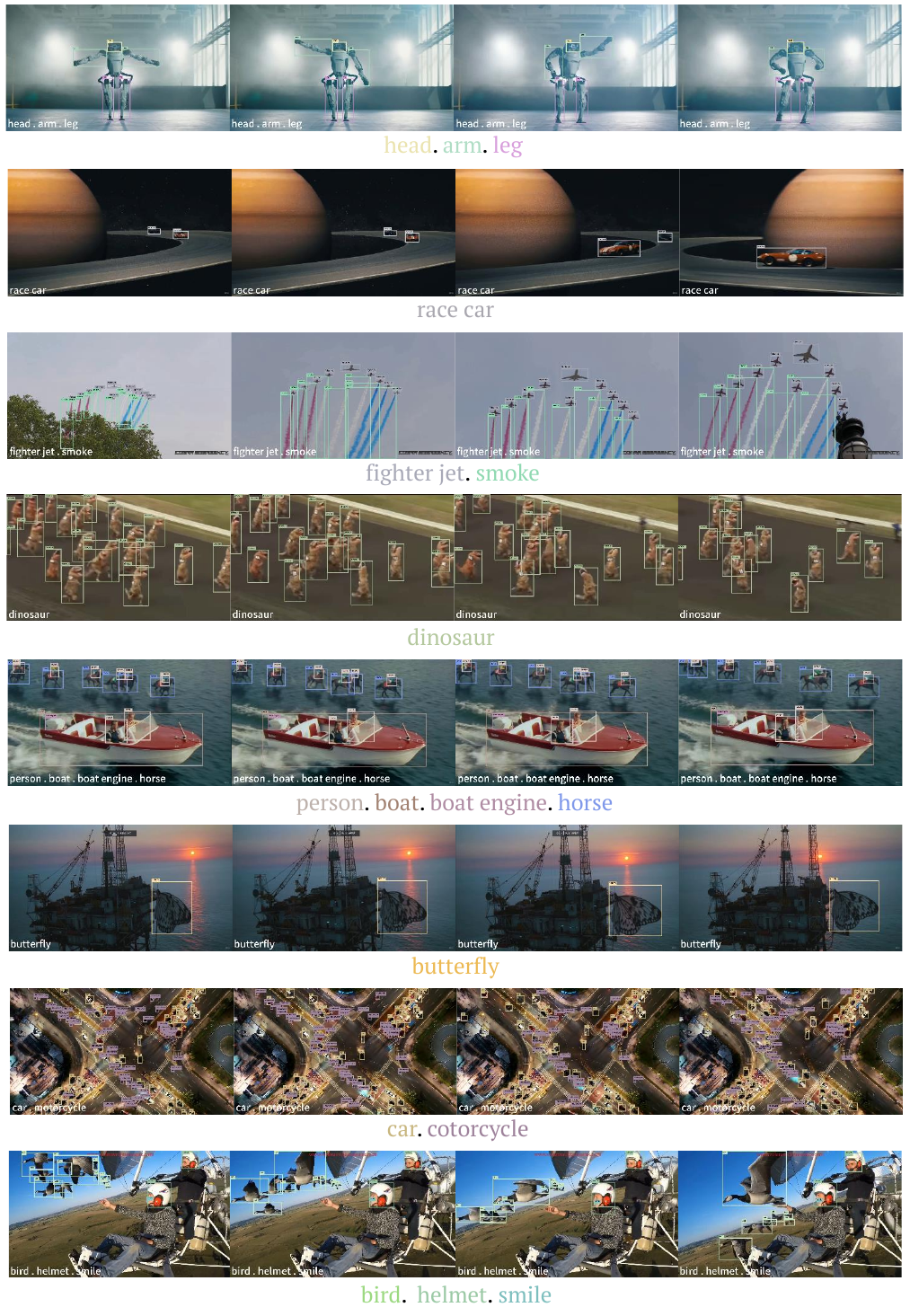}
 \caption{Model predictions on video object detection with Grounding DINO 1.5 Pro.
}
 \label{fig:video_vis}
\end{figure}

We present video detection results of Grounding DINO 1.5 Pro in Figure \ref{fig:video_vis}. We notice the model can produce consistent object bounding boxes in most cases. The videos are processed offline.

\subsection{Side-by-side Comparison}

\begin{figure}[ht!]
\centering
\includegraphics[width=\textwidth,keepaspectratio]{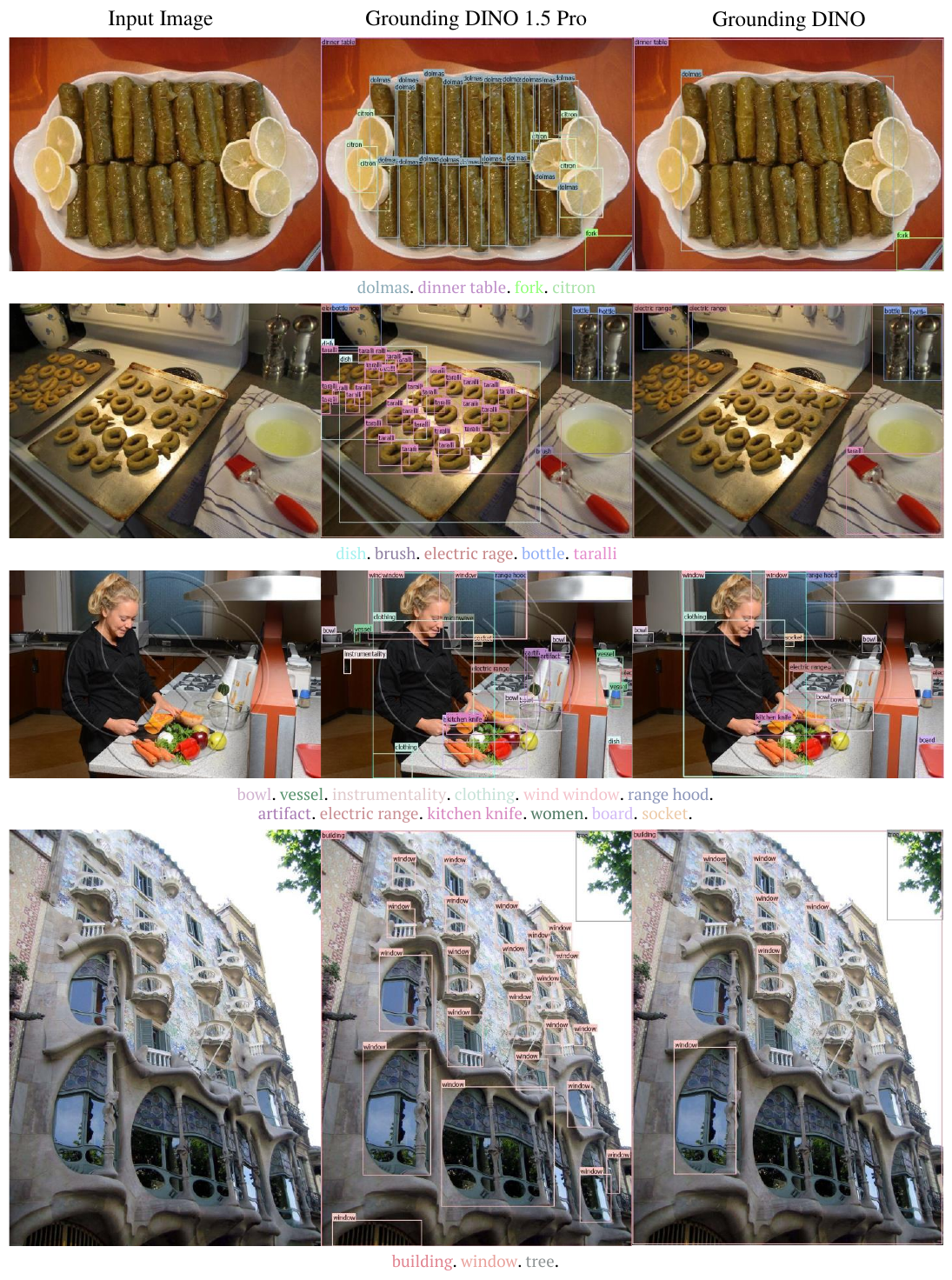}
 \caption{Side-by-side comparison between Grounding DINO 1.5 Pro and Grounding DINO~\cite{GroundingDINO} (part 1).
}
 \label{fig:side_by_side_compare_part1}
\end{figure}

\begin{figure}[ht!]
\centering
\includegraphics[width=\textwidth,keepaspectratio]{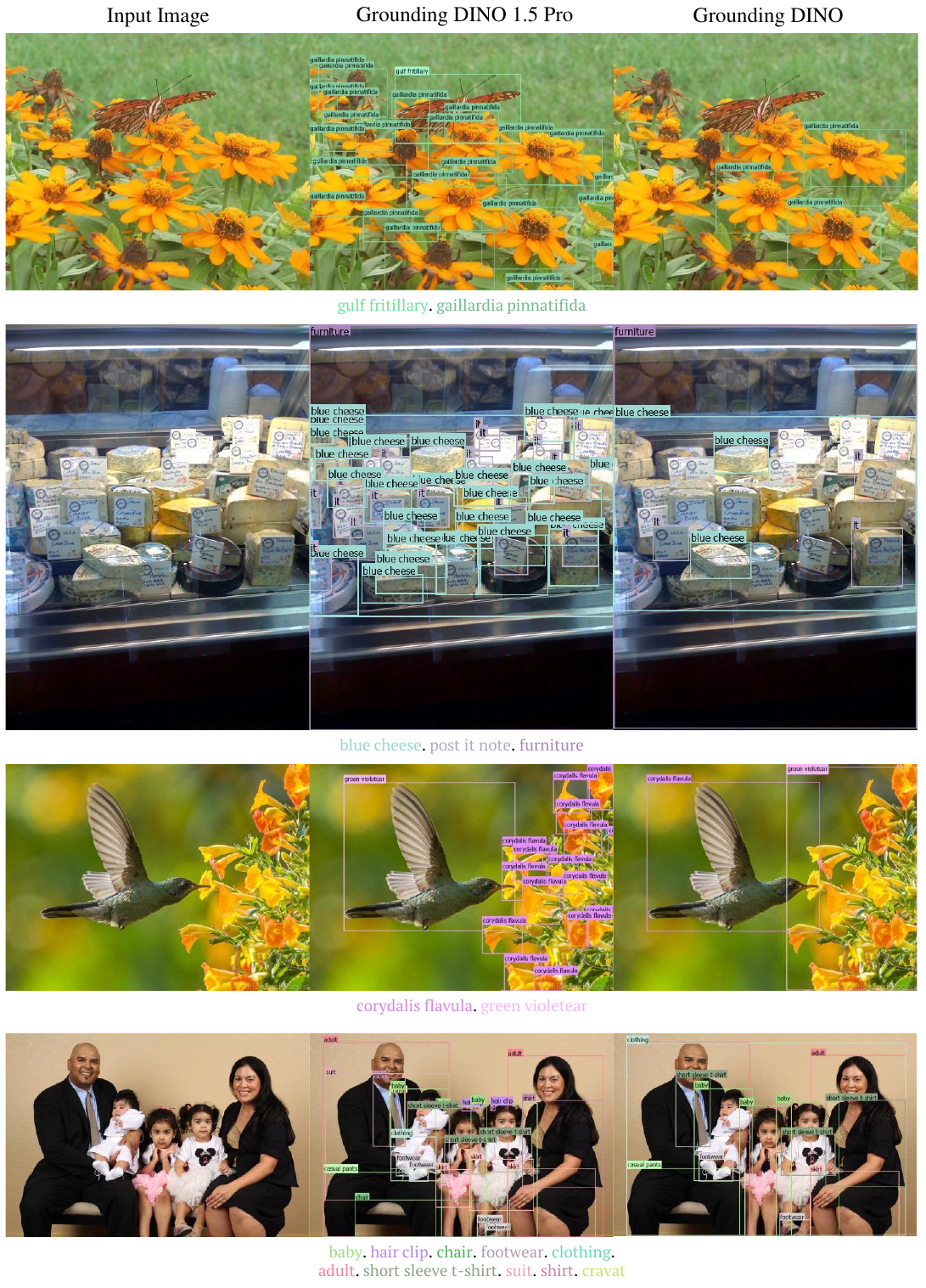}
 \caption{Side-by-side comparison between Grounding DINO 1.5 Pro and Grounding DINO~\cite{GroundingDINO} (part 2).
}
 \label{fig:side_by_side_compare_part2}
\end{figure}

We present the side-by-side comparison between the results of Grounding DINO 1.5 Pro and Grounding DINO in Figure \ref{fig:side_by_side_compare_part1} and Figure \ref{fig:side_by_side_compare_part2}. The Grounding DINO 1.5 Pro model demonstrates superior performance over the Grounding DINO model in terms of dense scene detection, long-tailed object detection, and the accuracy of semantic understanding.

\begin{figure}[ht!]
\centering
\includegraphics[width=\textwidth,keepaspectratio]{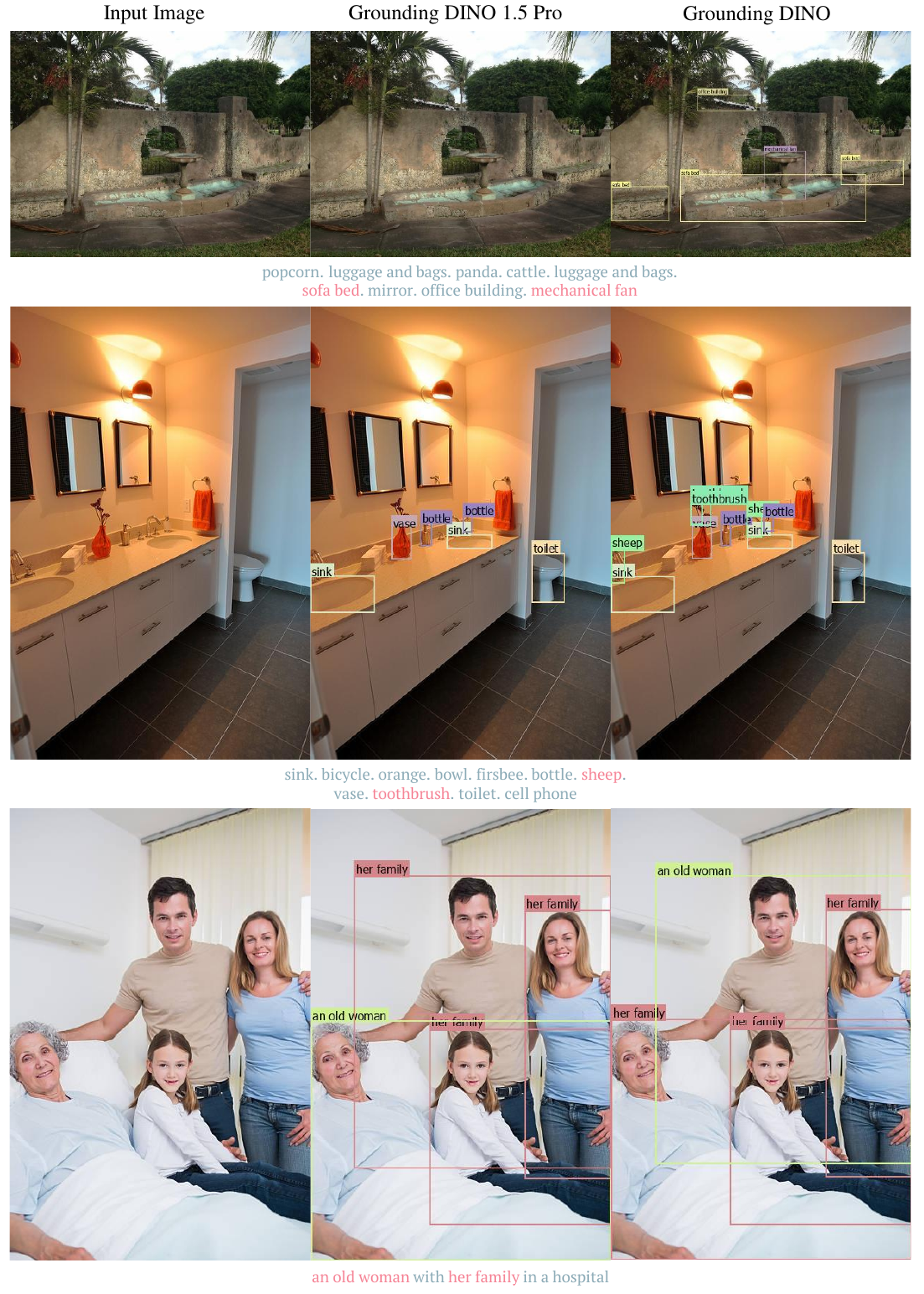}
 \caption{Side-by-side comparison between Grounding DINO 1.5 Pro and Grounding DINO~\cite{GroundingDINO} regarding object hallucinations. 
}
 \label{fig:vis_halluci}
\end{figure}

Moreover, we compare the object hallucinations of Grounding DINO 1.5 Pro and Grounding DINO, as shown in Figure \ref{fig:vis_halluci}. The results show that Grounding DINO 1.5 Pro has better accuracy and fewer object hallucinations. The last line in Figure \ref{fig:vis_halluci} demonstrates the better context understanding ability of our model.

\subsection{Advanced Object Detection on Edge Devices}

\begin{figure}[t!]
    \centering
    \includegraphics[width=\textwidth]{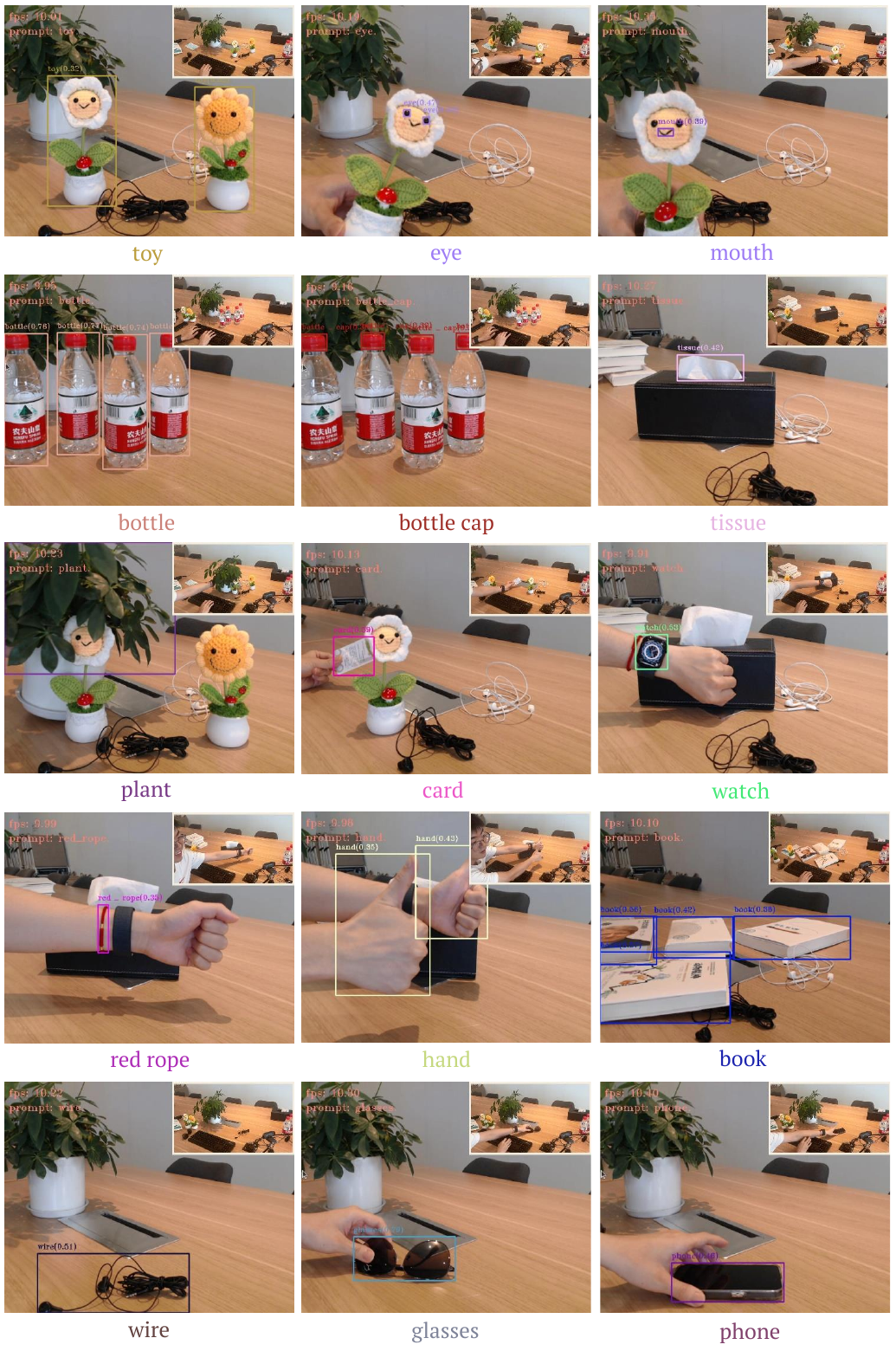}
    \caption{The visualization of Grounding DINO 1.5 Edge on NVIDIA Orin NX. The top left of the screen displays the FPS and prompts, while the top right shows a camera view of the recorded scene.}
    \label{fig:edge_vis}
\end{figure}

We present the practical, real-time application potential of Grounding DINO 1.5 Edge through a series of demonstrations in Figure \ref{fig:edge_vis}. The model's adept performance in office environments is particularly highlighted, offering significant utility for the field of robotics research.

\clearpage
\section{Conclusion}
This paper has presented Grounding DINO 1.5, a series of models to advance the field of open-set object detection. 
The flagship model, Grounding DINO 1.5 Pro, has established new records on the COCO and LVIS zero-shot benchmarks, signifying a major stride in detection accuracy and reliability. 
Moreover, the Grounding DINO 1.5 Edge model enables real-time object detection across various applications, further expanding the practical utility of the Grounding DINO 1.5 series.
\section{Contributions and Acknowledgments}
We would like to express our gratitude to everyone involved in the Grounding DINO 1.5 project. The contributions are as follows (in no particular order):

\vspace{-5pt}
\begin{itemize}[leftmargin=*]
    \item \textbf{Grounding DINO 1.5 Pro Model Design, Training Infra Development, Data Collection, Model Training and Model Evaluation}: Tianhe Ren, Qing Jiang, Shilong Liu, and Zhaoyang Zeng.
    \item \textbf{Grounding DINO 1.5 Edge Model Design, Training, Evaluation, and Runtime Optimization}: Wenlong Liu, Han Gao, Qing Jiang, Hongjie Huang, Zhengyu Ma, Xiaoke Jiang, and Yihao Chen.
    \item  \textbf{Grounding-20M Data Collection and Annotation Pipeline Construction}: Yihao Chen, Hao Zhang, Yuda Xiong, Tianhe Ren, Zhaoyang Zeng, Qing Jiang, Shilong Liu, and Peijun Tang.
    \item \textbf{Provide Great Insight and Technical Support}: Hao Zhang and Feng Li.
    \item \textbf{Grounding DINO 1.5 Edge Model Lead}: Kent Yu.
    \item \textbf{Overall Project Lead of Grounding DINO 1.5}: Lei Zhang.
\end{itemize}

We would also like to thank everyone involved in the Grounding DINO 1.5 demo support, including application lead Wei Liu, product manager Qin Liu and Xiaohui Wang, front-end developers Yuanhao Zhu, Ce Feng, and Jiongrong Fan, back-end developers Weiqiang Hu and Zhiqiang Li, UX designer Xinyi Ruan, tester Yinuo Chen, and Zijun Deng for helping with demo videos.

\clearpage
\section{Appendix}
\subsection{Detailed results on the ODinW benchmark}
\label{odinw_detailed_performance}

We report the detailed results of Grounding DINO 1.5 Pro on the ODinW35 benchmarks in Table~\ref{tab:odinw35_detailed_results}

\begin{table}[!h]
\setlength{\tabcolsep}{3pt}
\renewcommand{\arraystretch}{1.25}
\centering
\scriptsize
\resizebox{0.95\linewidth}{!}{
\begin{tabular}{l|cccc}
\toprule
Datasets & ODinW13  & ODinW35 &  Grounding DINO 1.5 Pro \tiny{(zero-shot)} & Grounding DINO 1.5 Pro \tiny{(fine-tuning)} \\
\midrule
AerialMaritimeDrone\_large & \checkmark & \checkmark  & 19.0 & 37.0 \\
AerialMaritimeDrone\_tiled &  &  \checkmark & 18.2 & 45.0 \\
AmericanSignLanguageLetters & & \checkmark & 13.7 & 83.3 \\
Aquarium & \checkmark  & \checkmark & 38.5 & 60.2 \\
BCCD & & \checkmark &  22.8 & 64.0 \\
ChessPieces & & \checkmark & 6.8 & 80.5 \\
CottontailRabbits & \checkmark & \checkmark & 65.7 & 75.1 \\
DroneControl & & \checkmark & 8.3 & 79.8 \\
EgoHands\_generic & \checkmark & \checkmark & 61.8 & 78.6 \\
EgoHands\_specific & & \checkmark & 16.7 & 78.7 \\
HardHatWorkers & & \checkmark & 20.5 & 46.4 \\
MaskWearing & & \checkmark & 16.7 & 63.2 \\
MountainDewCommercial & & \checkmark & 24.9 & 33.1 \\
NorthAmericaMushrooms & \checkmark & \checkmark & 82.1 & 89.2 \\
OxfordPets\_by\_breed & & \checkmark & 0.9 & 89.7 \\
OxfordPets\_by\_species & & \checkmark & 61.6 & 91.5 \\
PKLot & & \checkmark & 4.4 & 96.5 \\
Packages & \checkmark & \checkmark & 58.1 & 72.1 \\
PascalVOC & \checkmark & \checkmark &  67.1 & 77.6 \\
Raccoon & \checkmark & \checkmark & 72.5 & 81.8 \\
ShellfishOpenImages & \checkmark & \checkmark & 62.0 & 70.8 \\
ThermalCheetah & & \checkmark & 20.7 & 58.3 \\
UnoCards &  & \checkmark & 1.7 & 89.3 \\
VehiclesOpenImages & \checkmark  & \checkmark & 64.3 & 74.6 \\
WildfireSmoke &  & \checkmark & 28.9 & 57.5 \\
boggleBoards &  & \checkmark & 0.8 & 77.0 \\
brackishUnderwater & & \checkmark & 10.1 & 76.8 \\
dice &  & \checkmark & 0.6 & 79.5 \\
openPoetryVision & & \checkmark & 0.9 & 81.2 \\
pistols & \checkmark  & \checkmark &  71.9 & 77.6 \\
plantdoc & & \checkmark & 3.3 & 62.6 \\
pothole & \checkmark & \checkmark & 29.0 & 62.4 \\
selfdrivingCar &  & \checkmark & 7.4 & 53.1 \\
thermalDogsAndPeople & \checkmark & \checkmark &  71.4 & 84.0 \\
websiteScreenshots & & \checkmark & 2.3 & 41.6 \\
\midrule
ODinW13 Average AP & & & 58.7 & 72.4 \\
ODinW35 Average AP & & & 30.2 & 70.6 \\
\bottomrule
\end{tabular}
}
\caption{Detailed Zero-shot Results of Grounding DINO 1.5 Pro on the ODinW35 benchmark.}
\label{tab:odinw35_detailed_results}
\end{table}

\clearpage
\bibliography{main}

\begin{thebibliography}{10}

\bibitem{Cai_2023_ICCV}
Han Cai, Junyan Li, Muyan Hu, Chuang Gan, and Song Han.
\newblock Efficientvit: Lightweight multi-scale attention for high-resolution dense prediction.
\newblock In {\em Proceedings of the IEEE/CVF International Conference on Computer Vision (ICCV)}, pages 17302--17313, October 2023.

\bibitem{CC12M}
Soravit Changpinyo, Piyush Sharma, Nan Ding, and Radu Soricut.
\newblock {Conceptual 12M: Pushing Web-Scale Image-Text Pre-Training To Recognize Long-Tail Visual Concepts}.
\newblock {\em CVPR}, 2021.

\bibitem{cheng2024yolo}
Tianheng Cheng, Lin Song, Yixiao Ge, Wenyu Liu, Xinggang Wang, and Ying Shan.
\newblock {YOLO-World: Real-Time Open-Vocabulary Object Detection}.
\newblock {\em CVPR}, 2024.

\bibitem{FixedAP}
Achal Dave, Piotr Dollár, Deva Ramanan, Alexander Kirillov, and Ross Girshick.
\newblock {Evaluating Large-Vocabulary Object Detectors: The Devil is in the Details}.
\newblock {\em arXiv preprint arXiv:2102.01066}, 2022.

\bibitem{vit}
Alexey Dosovitskiy, Lucas Beyer, Alexander Kolesnikov, Dirk Weissenborn, Xiaohua Zhai, Thomas Unterthiner, Mostafa Dehghani, Matthias Minderer, Georg Heigold, Sylvain Gelly, Jakob Uszkoreit, and Neil Houlsby.
\newblock {An Image is Worth 16x16 Words: Transformers for Image Recognition at Scale}.
\newblock {\em ICLR}, 2020.

\bibitem{EVA-02}
Yuxin Fang, Quan Sun, Xinggang Wang, Tiejun Huang, Xinlong Wang, and Yue Cao.
\newblock {EVA-02: A Visual Representation for Neon Genesis}.
\newblock {\em arXiv preprint arXiv:2303.11331}, 2023.

\bibitem{LVIS}
Agrim Gupta, Piotr Dollár, and Ross Girshick.
\newblock {LVIS: A Dataset for Large Vocabulary Instance Segmentation}, 2019.

\bibitem{ResNet}
Kaiming He, Xiangyu Zhang, Shaoqing Ren, and Jian Sun.
\newblock {Deep Residual Learning for Image Recognition}.
\newblock {\em CVPR}, 2015.

\bibitem{TRex2}
Qing Jiang, Feng Li, Zhaoyang Zeng, Tianhe Ren, Shilong Liu, and Lei Zhang.
\newblock {T-Rex2: Towards Generic Object Detection via Text-Visual Prompt Synergy}.
\newblock {\em arXiv preprint arXiv:2403.14610}, 2024.

\bibitem{YOLOv8}
Glenn Jocher, Ayush Chaurasia, and Jing Qiu.
\newblock {Ultralytics YOLOv8}, January 2023.

\bibitem{kamath2021mdetr}
Aishwarya Kamath, Mannat Singh, Yann LeCun, Gabriel Synnaeve, Ishan Misra, and Nicolas Carion.
\newblock {MDETR - Modulated Detection for End-to-End Multi-Modal Understanding}.
\newblock {\em ICCV}, 2021.

\bibitem{SAM}
Alexander Kirillov, Eric Mintun, Nikhila Ravi, Hanzi Mao, Chloe Rolland, Laura Gustafson, Tete Xiao, Spencer Whitehead, Alexander~C. Berg, Wan-Yen Lo, Piotr Dollár, and Ross Girshick.
\newblock {Segment Anything}.
\newblock {\em ICCV}, 2023.

\bibitem{OpenImages}
Alina Kuznetsova, Hassan Rom, Neil Alldrin, Jasper Uijlings, Ivan Krasin, Jordi Pont-Tuset, Shahab Kamali, Stefan Popov, Matteo Malloci, Alexander Kolesnikov, Tom Duerig, and Vittorio Ferrari.
\newblock {The Open Images Dataset V4: Unified image classification, object detection, and visual relationship detection at scale}.
\newblock {\em IJCV}, 2020.

\bibitem{li2023visual}
Feng Li, Qing Jiang, Hao Zhang, Tianhe Ren, Shilong Liu, Xueyan Zou, Huaizhe Xu, Hongyang Li, Chunyuan Li, Jianwei Yang, et~al.
\newblock {Visual In-Context Prompting}.
\newblock {\em CVPR}, 2024.

\bibitem{LiteDETR}
Feng Li, Ailing Zeng, Shilong Liu, Hao Zhang, Hongyang Li, Lei Zhang, and Lionel~M. Ni.
\newblock {Lite DETR : An Interleaved Multi-Scale Encoder for Efficient DETR}.
\newblock {\em CVPR}, 2023.

\bibitem{GLIP}
Liunian~Harold Li*, Pengchuan Zhang*, Haotian Zhang*, Jianwei Yang, Chunyuan Li, Yiwu Zhong, Lijuan Wang, Lu~Yuan, Lei Zhang, Jenq-Neng Hwang, Kai-Wei Chang, and Jianfeng Gao.
\newblock {Grounded Language-Image Pre-training}.
\newblock {\em CVPR}, 2022.

\bibitem{COCO}
Tsung-Yi Lin, Michael Maire, Serge Belongie, Lubomir Bourdev, Ross Girshick, James Hays, Pietro Perona, Deva Ramanan, C.~Lawrence Zitnick, and Piotr Dollár.
\newblock {Microsoft COCO: Common Objects in Context}.
\newblock {\em ECCV}, 2014.

\bibitem{GroundingDINO}
Shilong Liu, Zhaoyang Zeng, Tianhe Ren, Feng Li, Hao Zhang, Jie Yang, Chunyuan Li, Jianwei Yang, Hang Su, Jun Zhu, et~al.
\newblock {Grounding DINO: Marrying DINO with Grounded Pre-Training for Open-Set Object Detection}.
\newblock {\em arXiv preprint arXiv:2303.05499}, 2023.

\bibitem{SwinTransformer}
Ze~Liu, Yutong Lin, Yue Cao, Han Hu, Yixuan Wei, Zheng Zhang, Stephen Lin, and Baining Guo.
\newblock {Swin Transformer: Hierarchical Vision Transformer using Shifted Windows}.
\newblock {\em ICCV}, 2021.

\bibitem{ConvNeXt}
Zhuang Liu, Hanzi Mao, Chao-Yuan Wu, Christoph Feichtenhofer, Trevor Darrell, and Saining Xie.
\newblock {A ConvNet for the 2020s}.
\newblock {\em CVPR}, 2022.

\bibitem{OWLViTV2}
Matthias Minderer, Alexey Gritsenko, and Neil Houlsby.
\newblock {Scaling Open-Vocabulary Object Detection}.
\newblock {\em NeurIPS}, 2023.

\bibitem{minderer2022simple}
Matthias Minderer, Alexey Gritsenko, Austin Stone, Maxim Neumann, Dirk Weissenborn, Alexey Dosovitskiy, Aravindh Mahendran, Anurag Arnab, Mostafa Dehghani, Zhuoran Shen, Xiao Wang, Xiaohua Zhai, Thomas Kipf, and Neil Houlsby.
\newblock {Simple Open-Vocabulary Object Detection with Vision Transformers}.
\newblock {\em ECCV}, 2022.

\bibitem{CLIP}
Alec Radford, Jong~Wook Kim, Chris Hallacy, Aditya Ramesh, Gabriel Goh, Sandhini Agarwal, Girish Sastry, Amanda Askell, Pamela Mishkin, Jack Clark, Gretchen Krueger, and Ilya Sutskever.
\newblock {Learning Transferable Visual Models From Natural Language Supervision}.
\newblock {\em ICML}, 2021.

\bibitem{APE}
Yunhang Shen, Chaoyou Fu, Peixian Chen, Mengdan Zhang, Ke~Li, Xing Sun, Yunsheng Wu, Shaohui Lin, and Rongrong Ji.
\newblock {Aligning and Prompting Everything All at Once for Universal Visual Perception}.
\newblock {\em CVPR}, 2024.

\bibitem{V3Det}
Jiaqi Wang, Pan Zhang, Tao Chu, Yuhang Cao, Yujie Zhou, Tong Wu, Bin Wang, Conghui He, and Dahua Lin.
\newblock {V3Det: Vast Vocabulary Visual Detection Dataset}.
\newblock {\em ICCV}, 2023.

\bibitem{wang2023detecting}
Zhenyu Wang, Yali Li, Xi~Chen, Ser-Nam Lim, Antonio Torralba, Hengshuang Zhao, and Shengjin Wang.
\newblock {Detecting Everything in the Open World: Towards Universal Object Detection}.
\newblock {\em CVPR}, 2023.

\bibitem{GLEE}
Junfeng Wu, Yi~Jiang, Qihao Liu, Zehuan Yuan, Xiang Bai, and Song Bai.
\newblock {General Object Foundation Model for Images and Videos at Scale}.
\newblock {\em CVPR}, 2024.

\bibitem{xu2024multi}
Yifan Xu, Mengdan Zhang, Chaoyou Fu, Peixian Chen, Xiaoshan Yang, Ke~Li, and Changsheng Xu.
\newblock {Multi-modal Queried Object Detection in the Wild}.
\newblock {\em NeurIPS}, 2023.

\bibitem{DetCLIPv2}
Lewei Yao, Jianhua Han, Xiaodan Liang, Dan Xu, Wei Zhang, Zhenguo Li, and Hang Xu.
\newblock {DetCLIPv2: Scalable Open-Vocabulary Object Detection Pre-training via Word-Region Alignment}.
\newblock {\em CVPR}, 2023.

\bibitem{DetCLIP}
Lewei Yao, Jianhua Han, Youpeng Wen, Xiaodan Liang, Dan Xu, Wei Zhang, Zhenguo Li, Chunjing Xu, and Hang Xu.
\newblock {DetCLIP: Dictionary-Enriched Visual-Concept Paralleled Pre-training for Open-world Detection}.
\newblock {\em NeurIPS}, 2022.

\bibitem{DetCLIPv3}
Lewei Yao, Renjie Pi, Jianhua Han, Xiaodan Liang, Hang Xu, Wei Zhang, Zhenguo Li, and Dan Xu.
\newblock {DetCLIPv3: Towards Versatile Generative Open-vocabulary Object Detection}.
\newblock {\em CVPR}, 2024.

\bibitem{SwinHuge}
Lu~Yuan, Dongdong Chen, Yi-Ling Chen, Noel Codella, Xiyang Dai, Jianfeng Gao, Houdong Hu, Xuedong Huang, Boxin Li, Chunyuan Li, Ce~Liu, Mengchen Liu, Zicheng Liu, Yumao Lu, Yu~Shi, Lijuan Wang, Jianfeng Wang, Bin Xiao, Zhen Xiao, Jianwei Yang, Michael Zeng, Luowei Zhou, and Pengchuan Zhang.
\newblock {Florence: A New Foundation Model for Computer Vision}.
\newblock {\em arXiv preprint arXiv:2111.11432}, 2021.

\bibitem{DINO}
Hao Zhang, Feng Li, Shilong Liu, Lei Zhang, Hang Su, Jun Zhu, Lionel~M. Ni, and Heung-Yeung Shum.
\newblock {DINO: DETR with Improved DeNoising Anchor Boxes for End-to-End Object Detection}.
\newblock {\em ICLR}, 2023.

\bibitem{zhang2023simple}
Hao Zhang, Feng Li, Xueyan Zou, Shilong Liu, Chunyuan Li, Jianwei Yang, and Lei Zhang.
\newblock {A Simple Framework for Open-Vocabulary Segmentation and Detection}.
\newblock In {\em Proceedings of the IEEE/CVF International Conference on Computer Vision}, pages 1020--1031, 2023.

\bibitem{GLIPv2}
Haotian Zhang, Pengchuan Zhang, Xiaowei Hu, Yen-Chun Chen, Liunian~Harold Li, Xiyang Dai, Lijuan Wang, Lu~Yuan, Jenq-Neng Hwang, and Jianfeng Gao.
\newblock {GLIPv2: Unifying Localization and Vision-Language Understanding}.
\newblock {\em NeurIPS}, 2022.

\bibitem{OmDet_Turbo}
Tiancheng Zhao, Peng Liu, Xuan He, Lu~Zhang, and Kyusong Lee.
\newblock {Real-time Transformer-based Open-Vocabulary Detection with Efficient Fusion Head}.
\newblock {\em arXiv preprint arXiv:2403.06892}, 2024.

\bibitem{zhao2023detrs}
Yian Zhao, Wenyu Lv, Shangliang Xu, Jinman Wei, Guanzhong Wang, Qingqing Dang, Yi~Liu, and Jie Chen.
\newblock {DETRs Beat YOLOs on Real-time Object Detection}.
\newblock {\em CVPR}, 2023.

\end{thebibliography}
\bibliographystyle{plain}





\end{document}